\documentclass[10pt,twocolumn,letterpaper]{article}

\usepackage{iccv}
\usepackage{times}
\usepackage{epsfig}
\usepackage{graphicx}
\usepackage{amsmath}
\usepackage{amssymb}
\usepackage{algorithm}
\usepackage{algorithmic}
\usepackage{multirow}
\usepackage{array}
\usepackage[nocompress]{cite}
\usepackage{subcaption}
\usepackage{arydshln} 
\usepackage{xr}
\usepackage{float}

\usepackage[pagebackref=true,breaklinks=true,letterpaper=true,colorlinks,bookmarks=false]{hyperref}
\usepackage[breaklinks=true,bookmarks=false]{hyperref}

\iccvfinalcopy 

\def\iccvPaperID{3201} 

\usepackage[capitalize]{cleveref}
\creflabelformat{equation}{#2#1#3}

\ificcvfinal\fi

\begin{document}

\title{Augmented Box Replay: Overcoming Foreground Shift \\ for Incremental Object Detection}
\author{Yuyang Liu\textsuperscript{1,2,3} \enspace Yang Cong\textsuperscript{4} \enspace Dipam Goswami\textsuperscript{5} \enspace Xialei Liu\textsuperscript{6} \enspace Joost van de Weijer\textsuperscript{5,7} \and\vspace{-13pt}\\ 
\textsuperscript{1}State Key Laboratory of Robotics, Shenyang Institute of Automation, Chinese Academy of Sciences \\
\textsuperscript{2} Institutes for Robotics and Intelligent Manufacturing, Chinese Academy of Sciences \\ 
\textsuperscript{3}University of Chinese Academy of Sciences 
\space\space \space\space 
\textsuperscript{4}South China University of Technology\\ 
\textsuperscript{5}Computer Vision Center, Barcelona \space\space\space\space
\textsuperscript{6}VCIP, CS, Nankai University\\
\textsuperscript{7}Department of Computer Science, Universitat Autònoma de Barcelona  \\ 
{\tt\small liuyuyang@sia.cn, congyang81@gmail.com \{dgoswami, joost\}@cvc.uab.es, xialei@nankai.edu.cn}
}

\maketitle
\ificcvfinal\thispagestyle{empty}\fi

\begin{abstract}
In incremental learning, replaying stored samples from previous tasks together with current task samples is one of the most efficient approaches to address catastrophic forgetting. However, unlike incremental classification, image replay has not been successfully applied to incremental object detection (IOD). In this paper, we identify the overlooked problem of foreground shift as the main reason for this. Foreground shift only occurs when replaying images of previous tasks and refers to the fact that their background might contain foreground objects of the current task. To overcome this problem, a novel and efficient Augmented Box Replay (ABR) method is developed that only stores and replays foreground objects and thereby circumvents the foreground shift problem. In addition, we propose an innovative Attentive RoI Distillation loss that uses spatial attention from region-of-interest (RoI) features to constrain current model to focus on the most important information from old model. ABR significantly reduces forgetting of previous classes while maintaining high plasticity in current classes. Moreover, it considerably reduces the storage requirements when compared to standard image replay. Comprehensive experiments on Pascal-VOC and COCO datasets support the state-of-the-art performance of our model~\footnote{Code is available at \href{https://github.com/YuyangSunshine/ABR_IOD.git}{https://github.com/YuyangSunshine/ABR\_IOD.git}}.
\end{abstract}

\section{Introduction}
\begin{figure}[t]
    \centering
    \includegraphics[width=0.9\linewidth,height=0.7\linewidth]{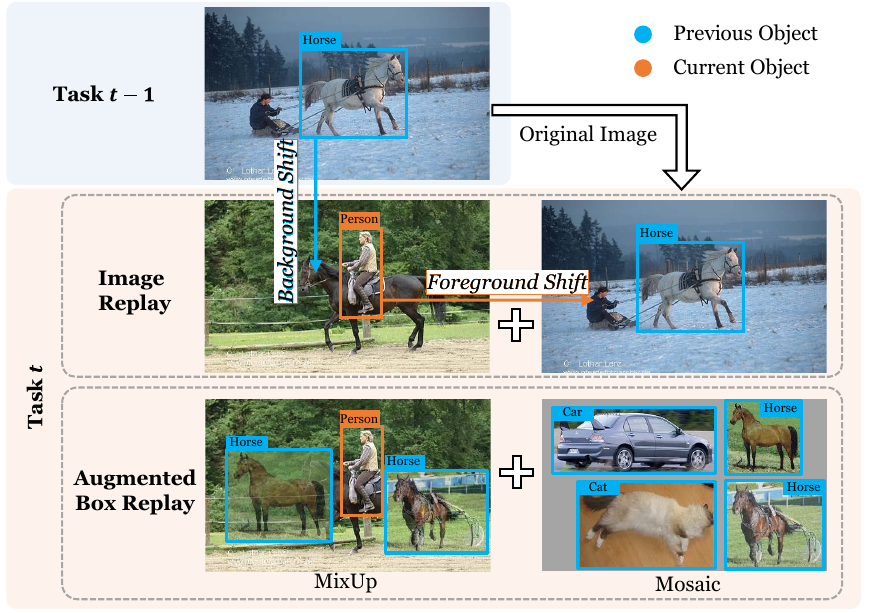}
    \vspace{-10pt}
    \caption{\textit{Background Shift} and \textit{Foreground Shift} for image replay settings. For each task, only the new classes are annotated while the other objects are considered as background (bkg). Moving from task $t-1$ to task $t$, the definition of the bkg changes, referred to as \textit{background shift}~\cite{cermelli2020modeling}. When current task samples are trained with exemplars from previous tasks, another critical problem-\textit{Foreground Shift} occurs due to varying annotations of \textit{new classes} between new samples (person as foreground) and exemplars (person as bkg) in the same task. Our augmented box replay method resolves these problems by mixing previous objects in the bkg of new images or fusing together for training.}
    \vspace{-15pt}
    \label{fig:introduction}
\end{figure}

\label{sec:intro}
The field of deep learning has witnessed remarkable progress recently, and state-of-the-art object detection models~\cite{carion2020end,ren2015faster,redmon2018yolov3,he2017mask,girshick2015fast} have been developed that performs exceptionally well on benchmark datasets. However, these models are typically designed to learn from data in a static manner, assuming that all object classes are available at once during training. In real-world scenarios, new object classes may emerge over time, making it necessary to update the model with new data. The inability to learn incrementally is a significant limitation for object detectors, particularly in cases of limited data storage capacity or data privacy concerns~\cite{de2021continual,masana2022class}. Therefore, developing incremental object detection (IOD) methods has become an essential and challenging task in real-world applications.

SOTA object detectors experience a phenomenon known as catastrophic forgetting~\cite{mccloskey1989catastrophic}, where their performance on previous classes degrades after learning new classes. This issue is commonly observed in incremental settings~\cite{de2021continual} and can be mitigated by balancing model stability (retaining previous information) and plasticity (learning new information without forgetting previous knowledge). 
While most studies in incremental learning are based on image classification~\cite{rebuffi2017icarl,li2017learning,kirkpatrick2017overcoming,aljundi2018memory}, recently it has been studied in the context of object detection~\cite{cermelli2022modeling,shmelkov2017incremental,hao2019end,chen2019new,peng2020faster} and semantic segmentation~\cite{douillard2021plop,zhang2022representation,goswami2023attribution}. A critical aspect in IOD is the background shift, also known as missing annotations~\cite{cermelli2022modeling, peng2020faster} which occurs due to the presence of multiple class objects in an image. Objects belonging to previous or future tasks in incremental object detection are often not annotated and assigned to the background class, as annotations are only available for classes in the current task. 

One of most efficient approaches in incremental classification is rehearsal-based strategy with storing images~\cite{rebuffi2017icarl,castro2018end}.
However, directly applying the replay images into IOD will cause the unlabelled objects of current classes in the replay images to be treated as background by the model. Consequently, the new objects will be background in replay images, while regarded as foreground in the new images. This leads to a contradiction between the foreground annotations in the exemplars and the current images as illustrated in~\cref{fig:introduction}. We refer to this problem as foreground shift which affects the plasticity of the current model.

To overcome the foreground shift for image replay in IOD, we propose a novel method called Augmented Box Replay (ABR). ABR uses mixup and mosaic box augmentation strategies to replay previous objects as an alternative to image replay for training in the current task. Compared to storing images in memory, ABR stores approximately four times as many object instances with the same storage requirements. To more effectively address catastrophic forgetting, we introduce a novel Attentive RoI Distillation loss that utilizes spatial attention from region-of-interest (RoI) features to align the most informative features of the previous and new models and correct the anchor position deviations of proposal pairs.

\begin{figure}[tbp]
\vspace{-10pt}
  \centering
  \begin{subfigure}[b]{0.23\textwidth}
    \includegraphics[width=\textwidth,height=0.8\textwidth]{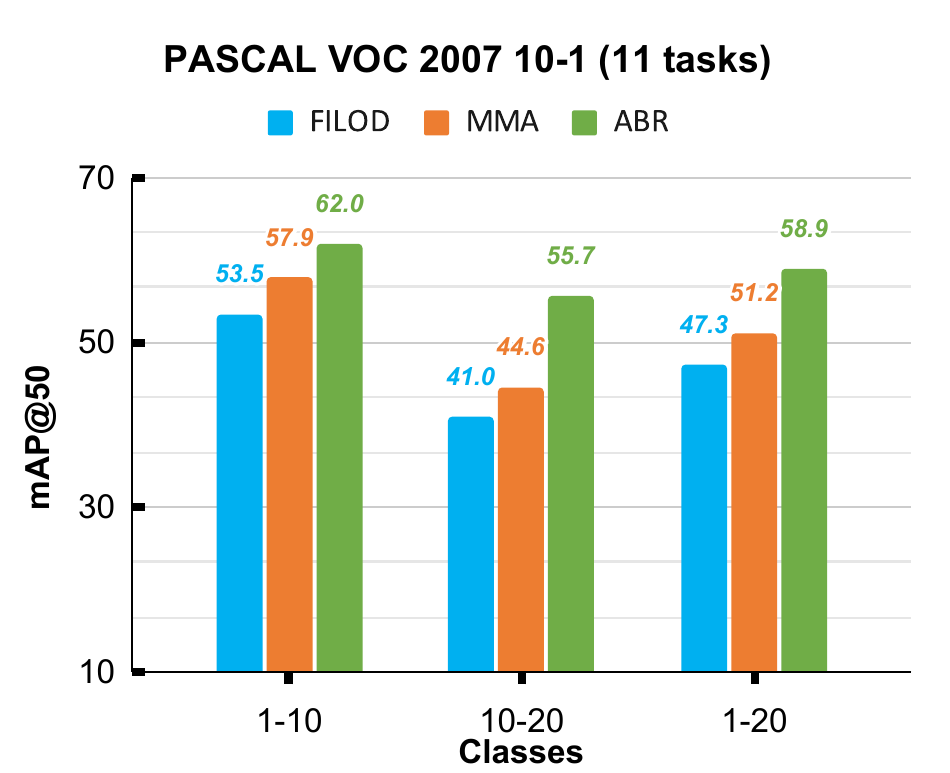}
  \end{subfigure}
  \vspace{-8pt}
  \begin{subfigure}[b]{0.23\textwidth}
    \includegraphics[width=\textwidth,height=0.8\textwidth]{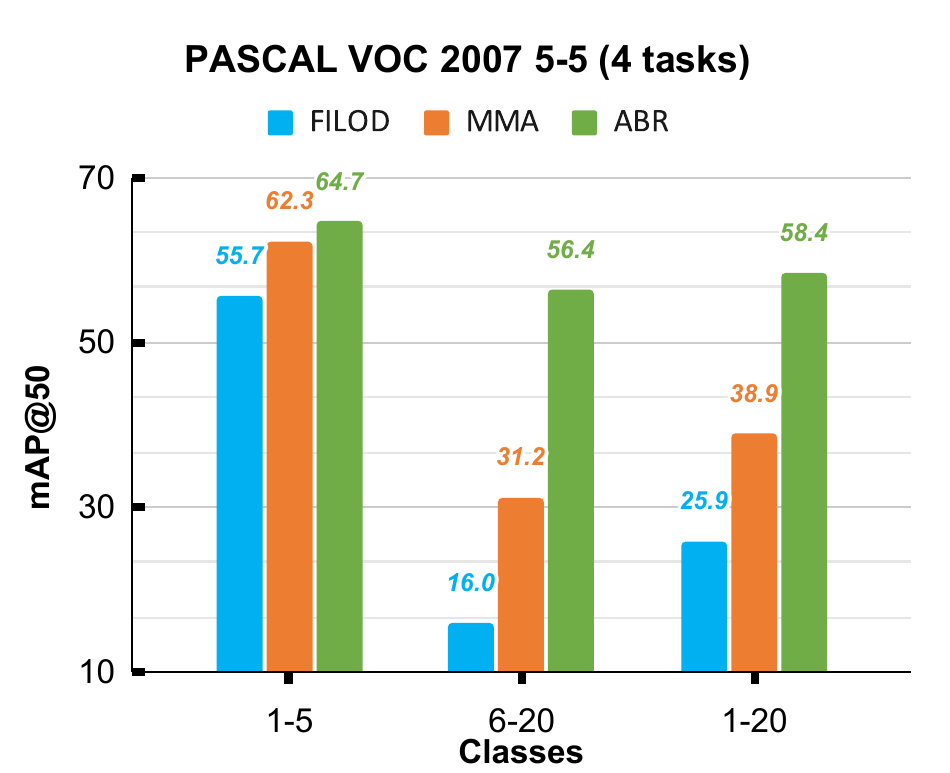}
  \end{subfigure}
  \caption{Our ABR method is especially well on the challenging longer sequences (10-1) and when starting with a small initial task (5-5). We compare here with state-of-the-art methods FILOD~\cite{peng2020faster} and MMA~\cite{cermelli2022modeling}.}
    \label{fig:longsteps}
    \vspace{-10pt}
\end{figure}

The proposed method is experimentally evaluated on Pascal-VOC and COCO datasets, and significantly outperforms SOTA methods in multiple settings. Our main contributions are three-fold: 
\vspace{-5pt}
\begin{itemize}
  \setlength\itemsep{0em}
    \item  This paper is the first to identify the critical foreground shift issue which has hampered the usage of replay methods for IOD. We propose Augmented Box Replay as a solution that reduces the memory requirements, eliminates the foreground shift, and improves the model stability and plasticity.
    \item We propose an innovative Attentive RoI Distillation loss to focus the current model on important location and feature information from the previous model and further reduce catastrophic forgetting.
    \item Our method outperforms state-of-the-art methods across multiple datasets and settings, showcasing its practicality and effectiveness. Especially, on the more challenging longer task sequences and the difficult scenario  with a small initial task, our method obtains significant performance gains (see~\cref{fig:longsteps}). 
\end{itemize}
\vspace{-5pt}

\vspace{-5pt}

\section{Related Work}
\label{sec:related}

\noindent\textbf{Object Detection:} Detector networks can be categorized into one-stage~\cite{tian2019fcos,tan2020efficientdet,liu2016ssd,carion2020end,redmon2018yolov3,lin2017focal} and two-stage~\cite{girshick2015fast,he2017mask,ren2015faster,lin2017feature} detectors. One-stage detectors which directly predict the output objects are comparatively faster while the two-stage detectors are generally superior in performance. The two-stage methods first extract regions of interests (RoIs) using a network~\cite{ren2015faster} and then obtain the final classification and regression outputs using a multi-layer network on the RoIs. Since these architectures perform poorly in incremental settings, we extend the two-stage Faster R-CNN~\cite{ren2015faster} network such that it can learn new object classes over time.


\noindent\textbf{Incremental Learning:} Class-incremental learning~\cite{masana2022class,de2021continual} and catastrophic forgetting~\cite{mccloskey1989catastrophic} has been explored extensively for image classification~\cite{castro2018end, li2017learning, rebuffi2017icarl} problems. The previous works can be categorized into rehearsal-based, parameter-isolation and regularization-based methods. Rehearsal-based methods store training samples~\cite{rebuffi2017icarl,castro2018end,li2022class} from previous tasks or generates training data~\cite{kemker2018fearnet,shin2017continual,wu2018memory}. Parameter-isolation methods~\cite{mallya2018packnet,mallya2018piggyback,yoon2018lifelong,liu2021l3doc} modify the initial network to accommodate new classes. Prior-focused regularization methods constrain learning on new classes and penalizing updating on weights~\cite{aljundi2018memory,kirkpatrick2017overcoming} or gradients~\cite{lopez2017gradient} while data regularization methods perform distillation~\cite{hinton2015distilling} between the intermediate features~\cite{douillard2020podnet,douillard2022dytox,li2017learning,hou2019learning} or attention maps~\cite{dhar2019learning} of the teacher model and the current student model to reduce forgetting. Other methods use embedding networks~\cite{Yu_2020_CVPR} or classifier drift correction~\cite{belouadah2019il2m} to address the changing class distributions. In our work, we focus on rehearsal-based and regularization-based methods.

\begin{figure*}[t]
    \centering
    \includegraphics[width=\linewidth]{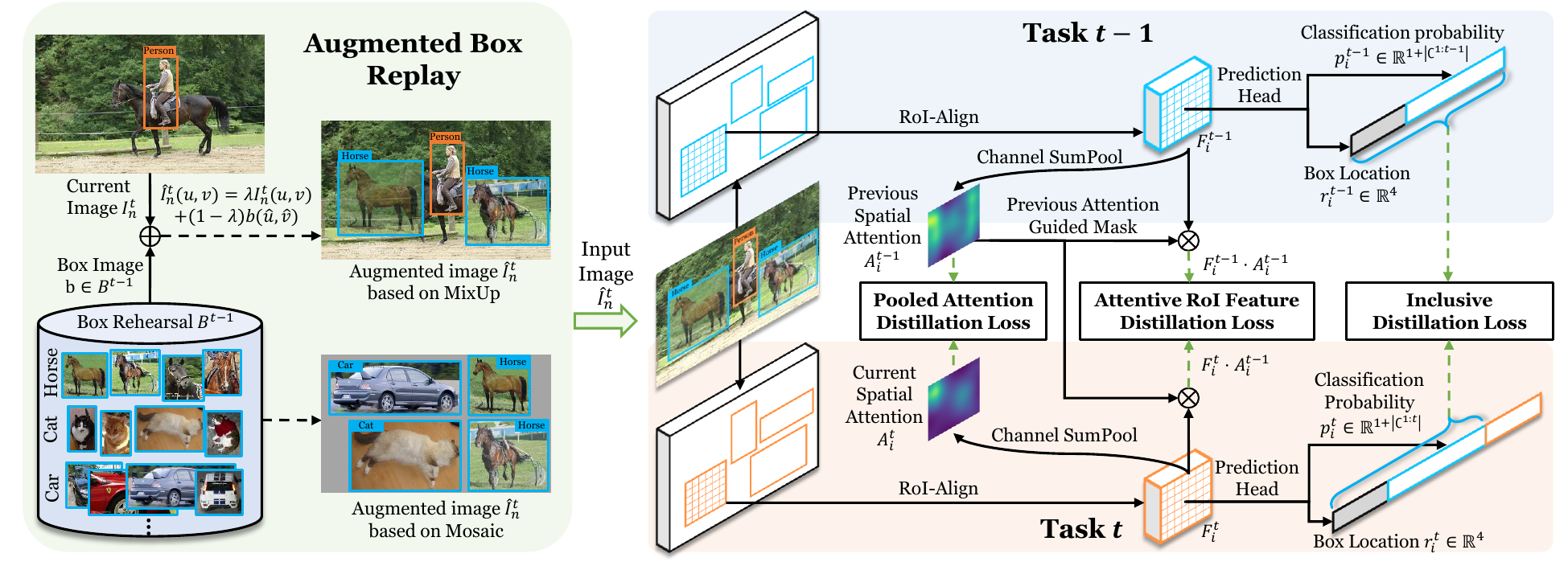}
    \vspace{-25pt}
    \caption{Illustration of our proposed framework, which highlights the key novelties of Augmented Box Replay (ABR) and Attentive RoI Distillation. ABR fuses prototype object $b$ from Box Rehearsal $B^{t-1}$ into the current image $I_n^t$ using mixup or mosaic. Attentive RoI Distillation uses pooled attention $A_i$ and masked features $F_i \cdot A_i^{t-1}$ to constrain the model to focus on important information from previous model. Inclusive Distillation Loss overcomes catastrophic forgetting based on ABR.}
    \label{fig:ABR}
    \vspace{-15pt}
\end{figure*}

\noindent\textbf{Incremental Object Detection:} 
Most of the recent works on incremental object detection use the Faster R-CNN~\cite{ren2015faster} architecture and performs distillation on the intermediate features~\cite{yang2022multi,liu2020multi,peng2020faster,hao2019end,cermelli2022modeling,zhou2020lifelong}, the region proposal network~\cite{cermelli2022modeling,peng2020faster,zhou2020lifelong} and head layers~\cite{feng2022overcoming}. 
Relatively few works~\cite{li2019rilod,peng2023diode,shieh2020continual} used one-stage architectures for incremental learning. Although the background shift issue was partially addressed in~\cite{zhou2020lifelong} by preventing previous class regions to be sampled as background but it was highlighted recently in~\cite{cermelli2022modeling,peng2020faster}.~\cite{cermelli2022modeling} proposed an unbiased classifier training loss and classifier distillation loss to explicitly tackle the background shift. EWC~\cite{kirkpatrick2017overcoming} has been adapted by~\cite{liu2020incdet} for object detection. 
While some methods replay images for finetuning~\cite{joseph2021towards,gupta2022ow} after training and for meta-learning~\cite{joseph2021incremental}, very few methods replay whole images~\cite{shieh2020continual} or stored feature representations~\cite{acharya2020rodeo} during training. For instance segmentation, \cite{ghiasi2021simple} explored copying random instances from one image to another.
Our work deals with bounding box replay methods to better address the challenges of IOD.


\section{Proposed Method}
\label{sec:method}

\subsection{Problem Formulation and Overview}
Object detection is primarily concerned with accurately identifying and localizing objects of interest within an image. 
Given a set of data $D=\{(I_n, Y_n)\}_{n=1}^{N}$, an ideal object detector $f_{\theta}(I_n)$ can predict a series of boxes $\hat{Y}_n$ corresponding to the groundtruth $Y_n$, where $Y_n = \{y_g=(u_g, v_g, w_g, h_g, c_g)\}_{g=1}^{G_n}$, with $(u_g, v_g)$ denoting the top-left corner coordinates of the bounding box and $(w_g, h_g)$ the width and height of the bounding box, and $c_i$ denotes the class for each of the $G_n$ bounding boxes. Therefore, $D$ has $K$=$\sum_{n=1}^{N} G_n$ groundtruth boxes totally.
This work focuses on two-stage detectors from the R-CNN family~\cite{ren2015faster,girshick2015fast,he2017mask} that typically consist of a CNN-based feature extractor, a Region Proposal Network (RPN), and a class-level classification and bounding box regression network (RCN). 

Incremental object detection aims to learn to detect objects in a sequence of $T$ tasks, where each task $D^t=\{(I^t_n, Y^t_n)\}_{n=1}^{N^t}$ corresponds to a new set of classes $\mathcal{C}^t$. The model should be able to detect objects in the new classes $\mathcal{C}^t$ while retaining the ability to detect objects in the previously seen classes $\mathcal{C}^{1:t-1}$ without catastrophic forgetting. However, unlike in the classification tasks where each input has a single label, $I^t_{n}$ may contain objects from both $\mathcal{C}^t$ and $\mathcal{C}^{1:t-1}$, and the annotations $Y^t_n$ only include the bounding boxes and class labels for $\mathcal{C}^t$. Therefore, $G^t_n\leq$ the number of the real annotations in IOD. 
The presence of unlabeled previous objects can lead to \textbf{\textit{Background Shift}} during training, where attention of the detector is biased towards the $\mathcal{C}^t$ and it fails to differentiate between the objects from $\mathcal{C}^t$ and $\mathcal{C}^{1:t-1}$. Moreover, misassociations can propagate over time, exacerbating catastrophic forgetting of previous classes. 

A straightforward way to a solution is using the original images from $D^{1:t-1}$, as shown in~\cref{fig:introduction}, which provides certain information for $\mathcal{C}^{1:t-1}$. However, the image replay method involves replaying original images from the previous training set during the current one, which can cause \textit{\textbf{Foreground Shift}} due to replay of unlabeled objects from $\mathcal{C}^t$. Thus, the new classes or the foreground in the current images are considered as the background in the replayed images which results in the model failing to generalize to new contexts. Additionally, storing the original images can result in significant memory overhead, since they include a lot of redundant information.

\subsection{Augmented Box Replay}

To mitigate the foreground shift problem, we propose an Augmented Box Replay (ABR) strategy that selects a subset of informative and representative box images from the previous task, along with a new set of boxes for the current task $t$. This method avoids replaying redundant information and optimally employs its storage for the relevant object regions. Specifically, ABR can replay these boxes in an augmented way, which helps the model retain its ability to detect previous objects in new contexts while improving its detection performance on the current task. \cref{fig:ABR} illustrates the pipeline of Augmented Box Replay strategy.

At the beginning, we involve a prototype box selection to choose the most representative boxes whose feature maps are close to the mean feature map after training of task $t-1$. The memory buffer is denoted as $B^{t-1}$, where the memory size $M^{t-1}$ of $B^{t-1}$ is limited. Therefore, the selection is an important factor that affects the performance. The final $B^{t-1}$ can focus on the most relevant information and avoid redundant or irrelevant information. Since box images are smaller than images, the storage cost is reduced, making it scalable to large datasets and complex models. See supplementary material for more details.

To leverage prototype boxes $B^{t-1}$ accumulated from the previous tasks in the current task $t$, we have designed two types of replay strategies: mixup box replay and mosaic box replay, inspired by~\cite{zhangmixup,bochkovskiy2020yolov4}. These strategies allow us to effectively transfer knowledge from past tasks to the current one and enhance the performance of the model.

\noindent  \textbf{MixUp box replay.} This method replays the box images of the previous class in the current data, placed in such a way that the previous box objects blend into the image more naturally, while ensuring that they have minimal overlap with the groundtruth bounding boxes of the new class. It involves assigning a random location in the current image $I_n^t$ to each box image $b \in B^{t-1}$ with size $(w_b, h_b)$, and then mixing it with $I_n^t$ to create a new image $\hat{I}_n^t$. More specifically, $\hat{I}_n^t$ is obtained by overlaying  $b$ onto $I_n^t$ at a location with a mixing coefficient $\lambda$. For each pixel location in $\hat{I}_n^t$, if $(u,v)$ is not inside the box, then the original pixel value of $I_n^t$ is retained. If $(u,v)$ is inside the box, the mixed pixel value is computed by:
\begin{equation}
\begin{aligned}
\hat{I}_n^t(u,v) = \begin{cases}
\lambda I_n^t(u,v)\quad &\multirow{2}*{$\text{if } \max\limits_{g\in G^n_t} y_g \cup b \leq th$}\\
+ (1-\lambda) b({\hat{u},\hat{v}}), \\
I_n^t(u,v) & \text{otherwise}
\end{cases},
\end{aligned}
\end{equation}
where $\lambda$ is values with the [0, 1] range and is sampled from the Beta distribution~\cite{zhangmixup}, $b(\hat{u},\hat{v})$ is the pixel value of the box image $b$ at location $(\hat{u}=u-w_b,\hat{v}=v-h_b)$, $y_g \cup b$ is the intersection over union (IOU) between each groundtruth annotation $y_g, \forall {g}\in G_n^t$ and the box image $b$, and $th$ is a threshold value. If the maximum IOU over union between the groundtruth annotations and the box image $b \leq th$, then the pixel value at $(u,v)$ in the new image $\hat{I}_n^t$ is a mixture of the original pixel value $I_n^t(u,v)$ and the corresponding pixel value in the box image $b$. Otherwise, the original pixel value $I_n^t(u,v)$ is retained. Note that at most two boxes are mixed up in a single image $I_n^t$ since the boxes are selected randomly and the overlap condition limits the number of boxes that can be mixed up in a single image.

\noindent  \textbf{Mosaic box replay.} This method involves dividing $I_n^t$ into a grid and randomly selecting a subset of cells. Each cell is then replaced with a box image $b$ from $B^{t-1}$, and the resulting image $\hat{I}_n^t$ is used for rehearsal. In the mosaic box replay strategy, a composite image is formed by combining four box images into a single mosaic image. To create the composite image, a random location is first selected as the center point of the mosaic image. Then, each of the four boxes is resized to a new size that is proportional to the size of the mosaic image, with the scaling factor $\mu$ randomly sampled from the range of [0.4, 0.6]. The resized boxes are arranged in the four quadrants of the mosaic image, and the remaining areas are filled with a fixed color value. 

In summary, the Augmented Box Replay offers several advantages for incremental learning in object detection: 1) \textbf{Information Richness:} ABR selects the most informative and representative boxes for rehearsal, which preserves the accuracy and diversity of learned model. 2) \textbf{Enhanced generalization:} ABR serves as an augmentation method which gives a different background context to both previous and new classes and thus improves the generalization of the model. 3) \textbf{Memory efficiency:} ABR replays only a small set of representative box images instead of the entire images, which significantly reduces the memory requirement. 4) \textbf{Adaptability:} ABR can easily be integrated with different object detection models to improve their performance.

\subsection{Attentive RoI Distillation}

\begin{table}[t]
\centering
\caption{Influence of different detector components in Faster-ILOD~\cite{peng2020faster} on VOC 10-10 setting. }
\renewcommand\arraystretch{1}
\small
\setlength{\tabcolsep}{4.5pt}
\vspace{-5pt}
\begin{tabular}{c:c:c:c|ccc}
\textbf{Frozen} & \textbf{Feature} & \textbf{RPN} & \textbf{RCN} & \multicolumn{3}{c}{\textbf{VOC (10-10)}}\\ 
\textbf{Backbone} & \textbf{Distil.} &\textbf{Distil.} & \textbf{Distil.} &\textbf{ 1-10 }&\textbf{ 11-20 }&\textbf{ 1-20 }\\
 \hline\hline
 &$\checkmark$   &$\checkmark$   &$\checkmark$   & 70.3 & 53.0 &61.7 \\ \hline
$\checkmark$ &  & $\checkmark$ & $\checkmark$  & {70.7} & 53.3 & 62.0 \\
 &   &$\checkmark$   &$\checkmark$   &70.6 & {53.7} & {62.2}  \\\hline
&$\checkmark$   &   &$\checkmark$   & 69.8 & 53.3 & 61.6\\\hline
&$\checkmark$   &$\checkmark$  &  & 8.2 & 62.7 & 35.5 \\
\end{tabular} 
\label{tab:infrpn}
\vspace{-10pt}
\end{table}
Distillation-based methods~\cite{shmelkov2017incremental,peng2020faster,cermelli2022modeling} are commonly used in IOD, aiming to transfer the knowledge of a model trained on a previous task (teacher) to a current model (student) while simultaneously learning the new task.
To further explore the impact of the distillation operation on the forgetting of each detector component, an ablation study is evaluated on the Faster-ILOD model~\cite{peng2020faster} as shown in~\cref{tab:infrpn}.
We can find that the feature extractor has a minimal effect on forgetting when either freezing the backbone or applying the feature distillation operation, and the presence or absence of the RPN component only has a 0.1\% effect on forgetting. However, removing the distillation operation of the prediction head (RCN) results in a 26.2\% drop in performance. 
Our obtained analysis and~\cite{verwimp2022re} together suggest that forgetting mainly occurs at the classification head. However, a limitation of RPN distillation lies in its focus solely on extracting RPN modules, which provide region proposals without considering features within each proposal. Consequently, the distilled model may overlook informative features within the proposals, leading to suboptimal performance.
To address this, we propose the Attentive RoI Distillation (ARD) loss, which allows the student model to selectively focus on the most important features from the teacher model by aligning the spatial attention maps and masked features of each proposal. Moreover, ARD supports more inclusive RoI features for the final prediction and helps to overcome the forgetting problem in the classification head. 

To enable the model to focus on the most informative parts of an image, we calculate a spatial attention map $A^t_i$ for each $F^t_i$, $\forall i \in P^t_n$, where $P^t_n$ is the number of proposals. The spatial attention map is obtained by raising the absolute value of each feature plane $F^t_{i,d}$ to a power $p$ (in the experiments, $p=2$) based on~\cite{zagoruyko2016paying} and summing them up:
\begin{equation}
A^t_i=\sum^{C}_{d=1}\left|F^t_{i,d}\right|^{p}, \quad p>0,
\end{equation}
Our method employs spatial attention maps from previous and current models to emphasize the most informative features and suppress the less informative ones. More superficially, the pooled attention distillation (PAD) loss is:
\begin{equation}
    \mathcal{L}_{PAD}=\left\|A_{i}^{t-1}-A_{i}^{t}\right\|,
\end{equation}
where $A_{i}^{t-1}$ and $A_{i}^{t}$ are the spatial attention maps for the $i^{th}$ proposal in the previous and current models, respectively. PAD can transfer knowledge from a previously trained model to a new one in a progressive learning setting.
The key difference with existing distillation methods in IOD is that here we explicitly distill the knowledge on the location of the relevant features (this is encoded in the attention map). 
Furthermore, ours applies the attentive distillation into the aligned bounding boxes, which contain the very relevant both location and feature information. Specifically, the Attentive RoI Feature Distillation (AFD) loss is employed:
\begin{equation}
\begin{aligned}
\mathcal{L}_{AFD}= &\dfrac{1}{P_n}\sum^{P_n}_{i=1}\left(F_{i}^{t-1}-F_{i}^{t}\right)^{2}A_{i}^{t-1},
\end{aligned}
\end{equation}
where $P_n^t$ is the number of proposals for $I_n^t$, $F_i^{t-1}$ and $F_i^{t}$ are the features extracted from the previous and new models, respectively. The squared difference $(F_{i}^{t-1}-F_{i}^{t})^2$ penalizes larger deviations between the previous and new features, which further encourages the new model to reproduce informative features from the previous model. By using the attention maps to weight the MSE term, AFD ensures new model to focus on reproducing the most important features from the previous model, while allowing for some flexibility in reproducing the less informative features. The overall ARD loss function is defined as: 
\begin{equation}
    \mathcal{L}_{ARD}=\mathcal{L}_{AFD}+\gamma \mathcal{L}_{PAD}
    \label{eq:ard}
\end{equation}
where $\gamma$ is a hyperparameter that controls the strength of the regularization. 

ARD loss not only aligns the features of each proposal but also has an effect on the position deviation of each anchor point. This spatial attention feature alignment reduces the impact of background shift caused by the imbalance between new and previous classes and promotes knowledge transfer from the previous model to the new model.

\subsection{Inclusive Loss with Background Constraint}
To avoid forgetting in classification head, we followed the unbiased classification and distillation losses proposed by~\cite{cermelli2020modeling,cermelli2022modeling}. However, due to our Augmented Box Replay strategy, the input image $\hat{I}_n^t$ contains many annotations about previous objects. This means that using unbiased losses directly in this situation is not feasible, as it would ignore the positive influence of the $B^{t-1}$ on the previous categories during the training phase. Therefore, we involve Inclusive Loss with Background Constraint to adapt the ABR based on the unbiased classification and distillation losses. In detail, the Inclusive Classification Loss is defined as follows:
\begin{equation}
\mathcal{L}_{IC}=\frac{1}{P_n^t} \sum_{i=1}^{P_n^t} c_{i}
\begin{cases}\log (p_{i}^b + \sum\limits_{c=1}^{\mathcal{C}^{1:t-1} } p_{i}^c), 
&  c_i = \mathcal{C}^b \\ 
\sum\limits_{c=1}^{\mathcal{C}^{1:t} } c_i \log p_{i}^c,  
&  c_i\in \mathcal{C}^{1:t}\end{cases}
\end{equation}
where $c_{i}$ is the label of proposal $i$, $p_{i}^b$ is the probability as $\mathcal{C}^{b}$, $p_{i}^c$ is the probability as $\mathcal{C}^{t}$. For positive RoI of $\mathcal{C}^{1:t}$ in ABR, the standard RCN loss based on cross-entropy is maintained. However, for negative RoI, the sum of the probabilities of $\mathcal{C}^{1:t-1}$ is treated as $\mathcal{C}^{b}$, ensuring that the model does not learn to predict $\mathcal{C}^{1:t-1}$ for unlabeled objects.

Moreover, the inclusive distillation (ID) loss maintains the performance of task $t-1$ by aligning the probabilities of the previous model for the background class with the probabilities of the new model for both $\mathcal{C}^{b}$ and $\mathcal{C}^{t}$. The training data for ABR includes grountruth annotations from box rehearsal, and the teacher model can detect previous objects. Therefore, we only need to focus on each proposal of $\mathcal{C}^{t}$:
\begin{equation}
\mathcal{L}_{ID} =  \frac{1}{\Omega}
\begin{cases}p_{i}^{b, t-1} \log ( p_{i}^{b,t} + \sum\limits_{c=1}^{\mathcal{C}^t } p_{i}^{c,t}), 
&  c_i = \mathcal{C}^b \\ 
\sum\limits_{c=1}^{\mathcal{C}^{t-1} } p_{i}^{c, t-1} \log (p_{i}^{c,t}),  
&  c_i\in \mathcal{C}^{1:t}\end{cases}
\end{equation}
where $\Omega = |\mathcal{C}^{1:t-1}|+1$ is the number of previous and background classes, $p_{i}^{b, t-1}$ and $p_{i}^{c, t-1}$ are the classification probabilities of the background class and previous classes in task $t-1$, respectively, $p_{i}^{b,t}$ and $p_{i}^{c,t}$ are the classification probabilities of the background class and new classes in task $t$, respectively, for the proposal $i$, $p_{i}^{c,t}$ is the classification probability of previous classes and new classes for the proposal $i$ in the current task $t$.

\begin{table*}[ht]
\centering
\caption{mAP@0.5\% results on settings with single increments on Pascal-VOC 2007. Best among columns in \textbf{bold} and second best among columns are \underline{underlined}. Methods with * use exemplars. ${\dagger}$: results from re-implementation.}
\small
\setlength{\tabcolsep}{7pt}
\vspace{-10pt}
\begin{tabular}{l||ccc|ccc|ccc|ccc}
& \multicolumn{3}{c|}{\textbf{19-1}}  & \multicolumn{3}{c|}{\textbf{15-5}}   & \multicolumn{3}{c|}{\textbf{10-10}} & \multicolumn{3}{c}{\textbf{5-15}} \\ 
\textbf{\#Method} & \textbf{1-19}    & \textbf{20}    & \textbf{1-20}     & \textbf{1-15}    & \textbf{16-20}    & \textbf{1-20}   & \textbf{1-10}    & \textbf{11-20}    & \textbf{1-20} & \textbf{1-5}    & \textbf{6-20}    & \textbf{1-20} \\ \hline\hline
Joint Training   & 70.1 & 75.7 & 74.3 & 76.4 &	67.8 &	74.3 & 75.5 &  73.0 & 74.3 & 70.1 & 75.7 & 74.3\\
Fine-tuning	                & 11.8 & 	64.7 &	14.4  &	15.9 &	54.2& 25.5 & 2.6 &	63.4 &	32.9 & 6.9 & 63.1 & 49.1 \\	\hline
ILOD (FasterRCNN)${\dagger}$~\cite{shmelkov2017incremental}	& 69.8  & \underline{64.5}	 & 69.6 &	72.5 &	58.5  &	68.9 & \underline{69.8} &	53.7 &	61.7  
 & 61.0 & 37.3 & 43.2\\
Faster ILOD${\dagger}$~\cite{peng2020faster} & \underline{70.9} & 63.2 & \underline{70.6}  & \textbf{73.1} & 57.3 &	69.2 & 70.3 & 53.0 & 61.7 & 62.0 & 37.1 & 43.3\\
PPAS~\cite{zhou2020lifelong}    & 70.5 & 	53.0 &	69.2  &	- & - & - & 63.5 &	60.0 &	61.8 & - & - & - \\
MVC	\cite{yang2022multi}	            & 70.2 & 	60.6 &	69.7 &	69.4 &	57.9&	66.5&  66.2 &	66.0 &	66.1 & - & - & - \\ 	
MMA${\dagger}$~\cite{cermelli2022modeling}	 & \underline{70.9} & 	62.9 & 70.5 &	72.7 & \underline{60.6} & \underline{69.7}  & \underline{69.8} & \underline{63.9} & \underline{66.8} & \textbf{66.8} & \underline{57.2} &\underline{59.6}\\ \hline
ORE*~\cite{joseph2021towards} 		        & 69.4 & 	60.1 &	68.9  &	71.8 &	58.7&	68.5&	 60.4 &	68.8 &	64.6 &	- & - & - \\
OW-DETR*~\cite{gupta2022ow}		    & 70.2 & 	62.0 &	69.8  &	72.2 &	59.8&	69.1	& 63.5 &	67.9 &	65.7 &	- & - & - \\
Meta-ILOD*~\cite{joseph2021incremental}	    & \underline{70.9} & 	57.6 &	70.2 &	71.7 &	55.9&	67.8	& 68.4 &	64.3 &	66.3 &	- & - & - \\  \hline
\textbf{ABR (Ours)} & \textbf{71.0} & \textbf{69.7} & \textbf{70.9} & \underline{73.0} & \textbf{65.1} & \textbf{71.0} & \textbf{71.2} & \textbf{72.8} & \textbf{72.0} &	\underline{64.7} & \textbf{71.0} & \textbf{69.4} \\ 
\end{tabular}
\label{tab:voc_exp_ss}
\end{table*}
\begin{table*}[ht]
\centering
\caption{mAP@0.5\% results on settings with multiple increments on Pascal-VOC 2007. Best among columns in \textbf{bold} and second best among columns are \underline{underlined}. ${\dagger}$: results from re-implementation.}
\small
\renewcommand\arraystretch{1}
\setlength{\tabcolsep}{4pt}
\vspace{-10pt}
\label{tab:voc_exps_ms}
\begin{tabular}{l||ccc|ccc|ccc|ccc|ccc}

 & \multicolumn{3}{c|}{\textbf{10-5 (3 tasks)}} &\multicolumn{3}{c|}{\textbf{5-5 (4 tasks)}}  & \multicolumn{3}{c|}{\textbf{10-2 (6 tasks)}} & \multicolumn{3}{c|}{\textbf{15-1 (6 tasks)}} & \multicolumn{3}{c}{\textbf{10-1 (11 tasks)}} \\
\textbf{\#Method}   & \textbf{1-10} & \textbf{11-20} & \textbf{1-20} & \textbf{1-5} & \textbf{6-20} & \textbf{1-20} & \textbf{1-10}  & \textbf{11-20}   & \textbf{1-20}  & \textbf{1-15}   & \textbf{16-20}   & \textbf{1-20}   & \textbf{1-10}   & \textbf{11-20}   & \textbf{1-20}  \\ \hline \hline
Joint Training  & 75.5 & 73.0 & 74.3  & 70.1 & 75.7 & 74.3  & 75.5 &  73.0 & 74.3  & 76.4 & 67.8 & 74.3   & 75.5 &  73.0 & 74.3  \\ 
Fine-tuning  & 5.3 & 30.6 & 18.0   & 0.5 & 18.3 & 13.8  & 3.79 & 13.6 & 8.7  & 0.0 & 10.47 & 5.3   & 0.0 & 5.1 & 2.55  \\\hline
ILOD (FasterRCNN)${\dagger}$~\cite{shmelkov2017incremental} & 67.2 & 59.4  & 63.3   &58.5 &15.6 &26.3  & 62.1   & 49.8    & 55.9   & 65.6   & 47.6    & 60.2   & 52.9   & 41.5    & 47.2  \\
Faster ILOD${\dagger}$~\cite{peng2020faster}  & \underline{68.3} & 57.9  & 63.1  &55.7 &16.0 &25.9  & 64.2   & 48.6    & 56.4   & 66.9   & 44.5    & 61.3   & 53.5   & 41.0    & 47.3  \\ 
MMA${\dagger}$~\cite{cermelli2022modeling} & 67.4 & \underline{60.5}  & \underline{64.0}   & \underline{62.3} &\underline{31.2} &\underline{38.9}  & \underline{65.7}   & \underline{52.5}  & \underline{59.1}   & \underline{67.2}   & \underline{47.8}    & \underline{62.3}   & \underline{57.9}   & \underline{44.6}  &  \underline{51.2}  \\ \hline
\textbf{ABR (Ours)} & \textbf{68.7} & \textbf{67.1} & \textbf{67.9}  & \textbf{64.7} & \textbf{56.4} & \textbf{58.4}  & \textbf{67.0} & \textbf{58.1} & \textbf{62.6} & \textbf{68.7} & \textbf{56.7} & \textbf{65.7} & \textbf{62.0} & \textbf{55.7} & \textbf{58.9} \\
\end{tabular}
\vspace{-10pt}
\end{table*}

\section{Experiments}
\label{sec:experiments}

\subsection{Experimental Settings}
\noindent\textbf{Datasets:} We evaluate the proposed method on two publicly available datasets namely PASCAL VOC 2007~\cite{everingham2009pascal} and MS COCO 2017~\cite{lin2014microsoft}. PASCAL VOC 2007 contains 20 object classes and 9,963 images, 50\% of which is used for training and validation and the remaining 50\% for testing following~\cite{everingham2009pascal}. MS COCO 2017, as a challenging dataset, has 80 different object classes and provides 83,000 images for training, 40,000 for validation and 41,000 for testing. 


\noindent\textbf{IOD Protocols:} Following previous works on this topic~\cite{shmelkov2017incremental,joseph2021incremental,cermelli2022modeling}, we obey the same experimental protocols. Each training task contains all images which have at least one bounding box from a new class. The annotations are available only for the new classes while the previous and future classes are not annotated. This setting is practical and can also have repetitions of images across tasks.

\noindent\textbf{Implementation Details:} Similar to~\cite{shmelkov2017incremental,cermelli2022modeling,joseph2021incremental,yang2022multi,liu2020multi,peng2020faster,hao2019end,chen2019new,zhou2020lifelong}, we use the Faster R-CNN~\cite{ren2015faster} architecture with a Resnet-50~\cite{he2016deep} backbone pretrained on ImageNet~\cite{deng2009imagenet}. 
We train the network with SGD optimizer, momentum of 0.9 and weight decay of $10^{-4}$. We use a learning rate of 5 $\times 10^{-3}$ for the initial task and 2 $\times 10^{-3}$ for the subsequent tasks. We used 15K iterations for 5 or 10 class increments in a task and 5K iterations when adding 1 or 2 new classes.
We set the memory size as 2,000 for all the experiments on PASCAL VOC 2007, 10,000 for 70-10 and 5,000 for 40-40 settings on MS COCO 2017 respectively. Our method uses a stack to store boxes, which are randomly selected and placed (while considering overlap criteria) during each iteration. 
To balance the number of old and new objects, we determine the 1:1:2 ratio for mixup, mosaic, and new images based on comparisons across different settings.

\noindent\textbf{Evaluation:} 
We evaluate the methods in terms of mean average precision at 0.5 IoU threshold for
PASCAL VOC 2007. For MS COCO 2017, we also report the mAP at different IoU ranging from 0.5 to 0.95 IoU (mAP@[50:95]), at 0.50 IoU (mAP@50) and at 0.75 IoU (mAP@75). 

\begin{table}[t]
\centering
\vspace{-5pt}
\caption{mAP results on MS COCO 2014 at different IoU, where the best among columns in \textbf{bold}.}
\vspace{-8pt}
\small
\setlength{\tabcolsep}{3.2pt}
\label{tab:coco_exps}
\begin{tabular}{l||ccc|ccc}
 & \multicolumn{3}{c|}{\textbf{40-40 mAP@}} &\multicolumn{3}{c}{\textbf{70-10 mAP@}}  \\
\textbf{\#Method}   & $\mathbf{[50:95]}$ & $\mathbf{50}$ & $\mathbf{75}$ & $\mathbf{[50:95]}$ & $\mathbf{50}$ & $\mathbf{75}$\\\hline\hline
Joint Training  & 35.9  & 60.5 & 38.0 & 35.9 & 60.5 & 38.0  \\ 
Fine-tuning  & 19.0  & 31.2 & 20.4 & 5.6 & 8.6 & 6.2\\ \hline
Faster ILOD~\cite{peng2020faster} & 20.6 & 40.1 & - & 21.3 & 39.9 & - \\
MMA~\cite{cermelli2022modeling} & 33.0 & 56.6 & 34.6 & 30.2 & 52.1 & 31.5 \\ \hline
\textbf{ABR (Ours)} & \textbf{34.5} & \textbf{57.8} & \textbf{35.2} & \textbf{31.1} & \textbf{52.9} & \textbf{32.7} \\
\end{tabular}
\vspace{-5pt}
\end{table}

\subsection{Quantitative Evaluation}
Following previous works~\cite{cermelli2022modeling,joseph2021incremental,peng2020faster,yang2022multi,zhou2020lifelong,shmelkov2017incremental}, we evaluate our method on settings with different number of initial classes and one or more incremental tasks. We compare our method with two baselines, the Fine-tuning when the model is trained with the data incrementally without any regularization or data replay, and the Joint training when the model is trained on the entire dataset with all the annotations. All results are obtained after training of the last task.

\vspace{-1em}

\begin{table*}[t]
\centering
\caption{Ablation study highlighting contribution from different components, where the best among columns in \textbf{bold}. }
\renewcommand\arraystretch{1}
\small
\setlength{\tabcolsep}{4.5pt}
\vspace{-8pt}

\begin{tabular}{c:c:c|c:c:c|c:c|ccc|cccc}

\textbf{ RCN} & \textbf{RPN} & \textbf{RoI} & \multicolumn{3}{c|}{\textbf{Selection}} & \multicolumn{2}{c|}{\textbf{AugmentedType}} & \multicolumn{3}{c|}{\textbf{VOC (10-10)}} & \multicolumn{4}{c}{\textbf{VOC (10-5)}}\\ 
$\mathcal{L}_{IC}$,$\mathcal{L}_{ID}$ & \textbf{Distil.} & $\mathcal{L}_{ARD}$ & \textbf{PBS} & \textbf{Herding} & \textbf{Random} & \textbf{Mixup} & \textbf{Mosaic} &\textbf{ 1-10 }&\textbf{ 11-20 }&\textbf{ 1-20 }& \textbf{1-10} & \textbf{11-15} & \textbf{16-20} & \textbf{1-20} \\  
 \hline\hline
 $\checkmark$ &   &   &   &   &  &  &  &   43.5 &   {75.9} & 59.4 &65.1 &31.3 &59.8 &55.3 \\
$\checkmark$ & $\checkmark$ &  &  &  &   & & & 45.2 & 75.6 & 60.4  & 67.1 &30.5 &59.3 & 55.9\\
$\checkmark$ &   & $\checkmark$  &  &  &   &  &  & 47.9 &  \textbf{76.2} & 62.0 &67.0 & 35.6 & 58.4 & 57.0 \\ \hline
$\checkmark$ &   & $\checkmark$  & $\checkmark$ & &    & $\checkmark$ &   & 68.9 & 72.6 & 70.7 & {67.4} &72.8 &\textbf{63.5} & {67.7} \\ 
$\checkmark$ &   & $\checkmark$  & $\checkmark$ &  &  &    & $\checkmark$ &   {70.6} & 71.2 &   {70.9} &67.0 &70.7 &61.8 &66.6 \\\hline
$\checkmark$ & $\checkmark$  &  &$\checkmark$  &  &  &   $\checkmark$ & $\checkmark$ & 69.7  & 72.4 & 71.0  & 67.4 & \textbf{72.9} & 61.1 & 67.2  \\
$\checkmark$ &   & $\checkmark$  &  & & $\checkmark$  & $\checkmark$ & $\checkmark$ &  68.7 & 71.5 &  70.1 &  67.0 & 71.2 & 62.8 & 67.0 \\
$\checkmark$ &   & $\checkmark$ & &$\checkmark$  &   &  $\checkmark$ & $\checkmark$ &  69.4 & 71.6 &  70.5 &  67.4 &  72.3 &  61.1 &  67.2 \\
$\checkmark$ &   & $\checkmark$  &$\checkmark$  &  & &    $\checkmark$ & $\checkmark$ &  \textbf{71.2} & 72.8 &  \textbf{72.0} &  \textbf{68.7} &  {71.5} &  {62.8} &  \textbf{67.9} \\
\end{tabular} 
\label{tab:ablation1}
\end{table*}

\subsubsection{PASCAL VOC 2007}
For PASCAL-VOC 2007, we perform our experiments on 19-1, 15-5, 10-10 and 5-15 single incremental task settings adding 1, 5, 10, 15 classes respectively. For multi-step incremental settings, we evaluate on 10-5, 5-5, 10-2, 15-1 and 10-1 settings where we add 5, 5, 2, 1 and 1 classes respectively at every step till all the 20 classes are seen.

\vspace{2 pt}

\noindent\textbf{Single-step increments:}
We benchmark our ABR method against the existing methods on~\cref{tab:voc_exp_ss}. We notice that Fine-tuning suffers from catastrophic forgetting across all settings. 
ABR outperforms all other methods across all the settings, significantly improving over MMA on the new classes by 4.5 mAP on 15-5, 8.9 mAP on 10-10 and 9.8 mAP on 5-15. We argue that the enhanced stability and plasticity is due to the augmented box replay of previous classes and our effective attention distillation. Our improvements over the methods storing exemplars~\cite{gupta2022ow,joseph2021incremental,joseph2021towards} confirm the importance of the box replay for IOD. 

\vspace{2 pt}

\noindent\textbf{Multi-step increments:}
The catastrophic forgetting and the background shift problem is more crucial on the longer incremental settings as seen in the performance from~\cref{tab:voc_exps_ms}. Fine-tuning suffers from almost complete forgetting on the initial classes. 
ABR improves over the closest competitor MMA by 3.9 mAP on 10-5, 3.5 mAP on 10-2, 3.4 mAP on 15-1 and 7.7 mAP on the longest and most challenging setting 10-1. It is interesting to observe that most methods struggle on the 5-5 setting with only 5 initial classes while ABR improves over MMA by 19.5 mAP. 
This implies that the existing methods require more classes in the initial task to achieve better generalization and thus, fails to adapt to new classes when the first task has lesser classes in 5-5 setting. 
On the most difficult setting of 10-1 with 10 increments, ABR outperforms MMA by 4.1 mAP on the previous classes and 11.1 mAP on the new classes. Note that for multiple increment settings, the improvement in the performance of incremental classes is not only due to better learning of new classes but also due to lesser forgetting of the intermediate task classes after moving to new tasks.

\subsubsection{MS COCO 2017} 
For MS COCO 2017, we perform experiments on 40-40 and 70-10 settings adding 40 and 10 classes respectively. As shown in~\cref{tab:coco_exps}, Fine-tuning suffers from catastrophic forgetting on both settings. While Faster ILOD and MMA has improved over Fine-tuning, our method improves average mAP@[50:95] over MMA by 1.5 on 40-40 setting and by 0.9 on 70-10 setting. These results signify lesser forgetting and better adaptation to new classes with our method.



\subsection{Analysis and Ablation Study}
We investigate the role of the network components, replay selection strategies, augmentation types in~\cref{tab:ablation1} on the VOC 10-10 and 10-5 settings. We take the baseline model with the RCN classification and distillation loss proposed by~\cite{cermelli2022modeling}. We show that our attentive RoI distillation improves over the RPN distillation used by~\cite{cermelli2022modeling,peng2020faster} owing to better exploitation of location and feature information of the RoIs. 
In replay strategies, we implemented the herding strategy~\cite{rebuffi2017icarl} for selecting boxes to replay. Our method improves 1\%$\sim$1.5\% mAP over the herding strategy. We can observe that our proposed prototype box selection can better capture more representative prototype samples for previous classes. Further, we add mixup and mosaic replay individually and observe that both strategies improve the performance on previous and new classes. The best performance is achieved when both mixup and mosaic replay are performed with the new images.

\begin{figure}[t]
    \centering
    \vspace{-5pt}
    \includegraphics[width=\linewidth]{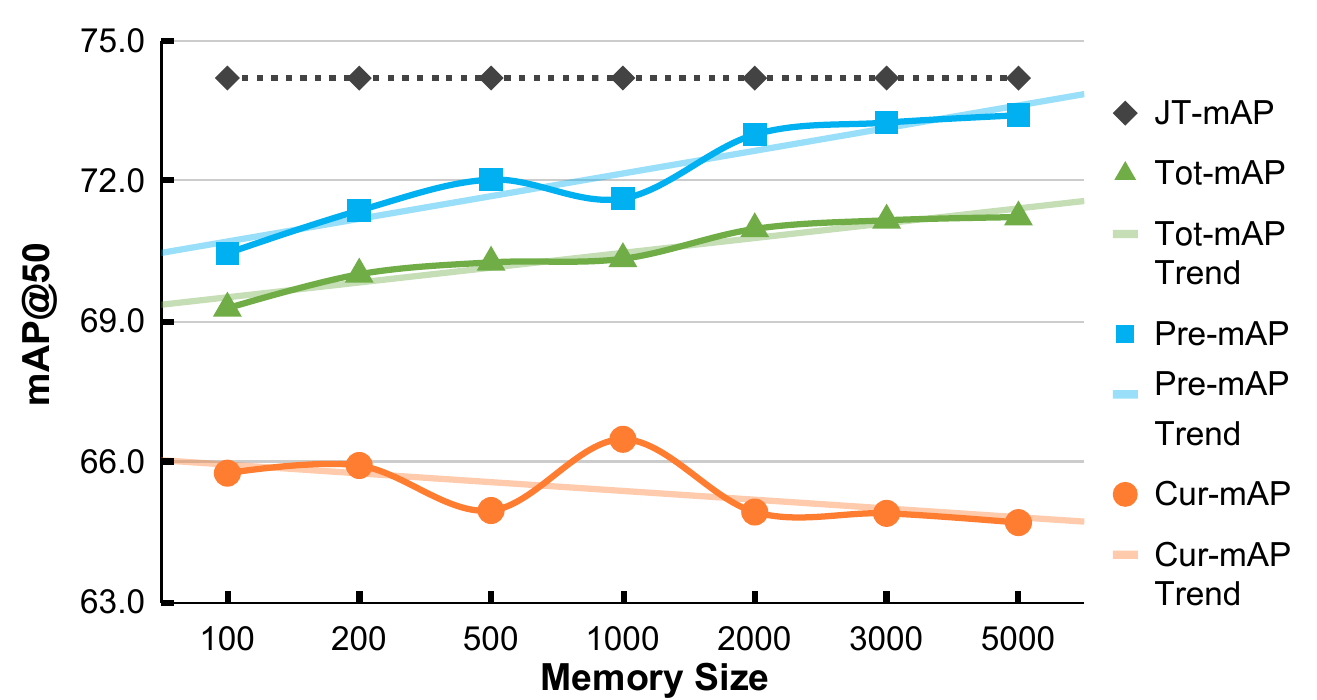}
    \vspace{-20pt}
    \caption{The average mAP@50 of previous, current and total classes in terms of different memory sizes at PASCAL VOC 2007 15-5 setting.}
    \label{fig:bf}
    \vspace{-10pt}
\end{figure}

We investigate the role of the memory size and train ABR with different memory size of previous class boxes. \cref{fig:bf} plots the mAP@50 results with increasing memory size. It is observed that the performance increases with increasing memory size or replay of more previous objects. It can be observed that after the memory size $>$ 2000, the growth rate of mAP tends to be more stable. Therefore, in the main experiments, we use a memory size of 2000.

\begin{table}[t]
\centering
\caption{Rehearsal alternative on Pascal VOC 2007 in mAP@50. All experiments are done in our proposed method with image replay or augmented box replay (ABR).}
\small
\renewcommand\arraystretch{1}
\setlength{\tabcolsep}{3pt}
\vspace{-8pt}
\begin{tabular}{c|c|c|c|ccc}
  &  &  &  & \multicolumn{3}{c}{\textbf{VOC (10-10)}} \\ 
\textbf{Type} & \textbf{Buffer Size} & \textbf{Objects} &\textbf{ Memory}$\downarrow$ & \textbf{1-10} & \textbf{11-20} & \textbf{1-20} \\ \hline\hline
  - & - & - & -  & 47.9  & 76.2 & 62.0 \\ \hline
 Image & 182 & 455 & 15.5Mb  & 70.2  & 62.2 & 66.2 \\ 
 Image & 800 & 2000  & 68Mb  & 71.6 & 57.9 & 64.7 \\ 
 ABR & 2000 & 2000 & 15.5Mb  & 71.2 & 72.3 & 72.0\\ 
\end{tabular}  
\label{tab:memory}
\vspace{-5pt}
\end{table}
\cref{tab:memory} presents a comparison between image replay and our proposed ABR method. The same number of objects ensures that the original information about the previous categories stored in the memory buffer is consistent, and the same storage space controls practicality in real-world applications. As shown in \cref{tab:memory}, despite having the same number of objects, image replay performs worse than augmented box replay in recognizing new classes. This confirms that replaying original images can lead to foreground shift and limit the adaptation of new classes. On the other hand, our memory buffer contains about 4 times as many original objects for previous classes as image replay.


\subsection{Visualization}
\begin{figure}[htbp]
  \centering
  \begin{subfigure}[b]{0.23\textwidth}
    \includegraphics[width=\textwidth,height=0.6\textwidth]{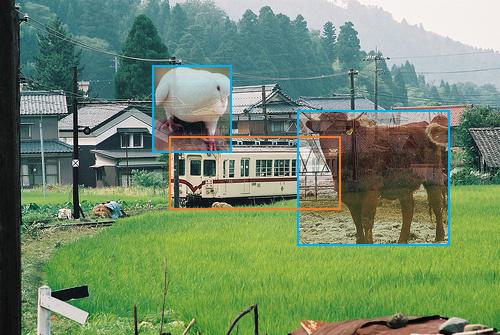}
    \includegraphics[width=\textwidth,height=0.6\textwidth]{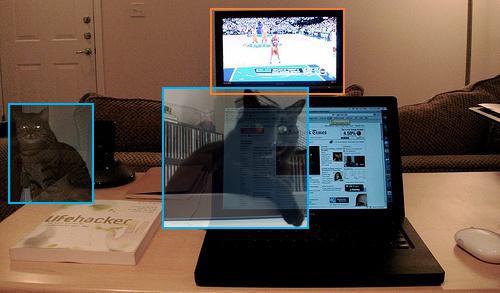}
  \end{subfigure}
  \begin{subfigure}[b]{0.23\textwidth}
    \includegraphics[width=\textwidth,height=0.6\textwidth]{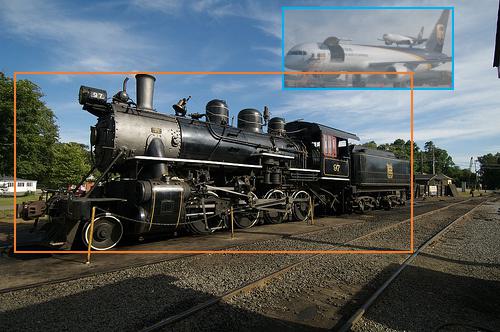}
    \includegraphics[width=\textwidth,height=0.6\textwidth]{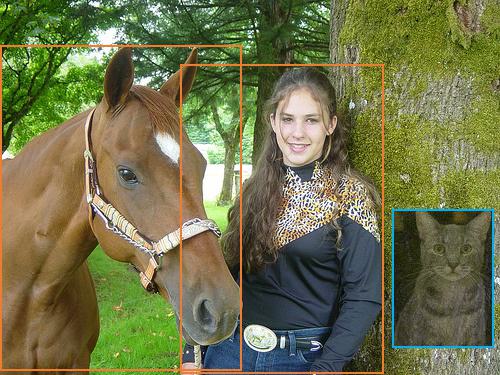}
  \end{subfigure}
  \caption{Examples of images generated by mixup augmentation for 10-10 setting on PASCAL VOC 2007. Blue boxes represent previous classes which are replayed in the background of new images. Orange boxes represent the ground truth annotations of current classes.}
    \label{fig:mixup}
    \vspace{-10pt}
\end{figure}

\cref{fig:mixup} shows some examples of images generated by mixup replay in VOC 10-10 setting. It can be seen intuitively that the mixup strategy makes the box reasonably integrated into the new images and minimizes the occlusion with the new objects. In addition, the background information compared to the new objects is greatly enriched.
The inference results are available in supplementary material.

\vspace{-5pt}

\section{Conclusion}
In this paper, we studied the experience replay method for incremental object detection problem and introduced the critical issue of foreground drift during old image replay. We hypothesize that the foreground drift is the reason that replay methods, which are dominant in incremental learning for image classification, have been little studied for IOD.

To tackle this problem, our proposed method ABR stores bounding boxes from old classes and replays them with new images using mixup and mosaic augmentation strategies. This overcomes the foreground drift situation since only the old classes are stored and replayed and not the unlabeled new classes from old images. In addition to box replay, the proposed attentive RoI distillation uses both the location and feature information for the RoIs extracted from the RPN and enables retention of meaningful knowledge of old classes. Further, our method reduces the memory overhead significantly. We demonstrate that ABR outperforms existing methods across all settings on representative datasets.

This work lays the foundation for bounding box replay instead of the traditional image or feature replay methods for object detection tasks. 
Future research should explore the implications of the foreground shift in incremental semantic segmentation and extend our approach to popular transformer methods~\cite{liu2023continual}.

\small{
\noindent\textbf{Acknowledgement.} This work is supported by National Natural Science Foundation of China (Grant No. 62127807, 62206135). We acknowledge projects TED2021-132513B-I00 and PID2022-143257NB-I00, financed by MCIN/AEI/10.13039/501100011033 and FSE+ and the Generalitat de Catalunya CERCA Program.}

{\small
\bibliographystyle{ieee_fullname}
\bibliography{egbib}
}

\clearpage
\twocolumn[
\begin{@twocolumnfalse}
    \centering{\Large\textbf{Supplementary Materials: Augmented Box Replay: \\ Overcoming Foreground Shift for Incremental Object Detection}\vspace{20pt}\\
    \large{Yuyang Liu\textsuperscript{1,2,3} \enspace Yang Cong\textsuperscript{4} \enspace Dipam Goswami\textsuperscript{5} \enspace Xialei Liu\textsuperscript{6} \enspace Joost van de Weijer\textsuperscript{5,7}\\
    \textsuperscript{1}State Key Laboratory of Robotics, Shenyang Institute of Automation, Chinese Academy of Sciences \\
    \textsuperscript{2} Institutes for Robotics and Intelligent Manufacturing, Chinese Academy of Sciences \\ 
    \textsuperscript{3}University of Chinese Academy of Sciences 
    \space\space \space\space 
    \textsuperscript{4}South China University of Technology\\ 
    \textsuperscript{5}Computer Vision Center, Barcelona\space\space\space\space 
    \textsuperscript{6}VCIP, CS, Nankai University\\
    \textsuperscript{7}Department of Computer Science, Universitat Autònoma de Barcelona\\}
    {\tt\small liuyuyang@sia.cn, congyang81@gmail.com, \{dgoswami, joost\}@cvc.uab.es, xialei@nankai.edu.cn}\vspace{25pt}}
\end{@twocolumnfalse}
]

\ificcvfinal\thispagestyle{empty}\fi
\appendix \label{sec:appendix}
\setcounter{table}{6}
\setcounter{figure}{5}
\setcounter{equation}{7}

\section{Additional Methods}
\subsection{Prototype Box Selection}\label{sec:pbs}

This method involves selecting the most representative boxes, as prototypes, from the current training data, which are then replayed along with the future training data. The memory buffer is commonly denoted as $B^t$, where $t$ represents the current task and the size $M$ of $B^t$ is limited. Therefore, the selection is an important factor that affects the performance. We employ a frozen trained model to generate the Region of Interest (RoI)-Aligned feature maps $\{F^{t}_g \in \mathbb{R}^{C\times S \times S}\}_{g=1}^{G^t_n}$ for $G^t_n$ groundtruth boxes in the current task $t$, where $C$ is the number of feature planes and $S$ is the spatial dimension. Then, a prototype feature map $\hat{F}^{t}_{c}$ for each class $c \in \mathcal{C}^{t}$ can be computed by:
\begin{equation}
\hat{F}^{t}_{c}=\frac{1}{|{F}^{t}_{c}|}\sum_{g=1}^{G^t_n} F^{t}_{g},\quad \forall c_g = c,
\label{eq:profm}
\end{equation}
The distance between each feature map $F^{t}_{g}$ and the prototype feature map $\hat{F}^{t}_{c}$ for class $c$ is computed using the Euclidean distance:
\begin{equation}
d(F^{t}_{g}, \hat{F}^{t}_{c}) = \sqrt{\sum (F^{t}_{g}-\hat{F}^{t}_{c})^2},\quad \forall c_g = c,
\label{eq:dis}
\end{equation}
Then we sort $\{d(F^{t}_{g}, \hat{F}^{t}_{c}),\forall c_g = c\}_{g=1}^{G^t_n}$ in ascending order, and select the top $M_c=\frac{M}{|\mathcal{C}^{1:t}|}$ boxes for that class to form the box buffer $B^{t}_{c}$. The final $B^{t}$ can focus on the most relevant information for each task and avoid redundant or irrelevant information, as shown in~\cref{alg:1}.

Additionally, since boxes are typically smaller than whole images, the computational cost of training and rehearsal can be reduced, making the approach more scalable to large datasets and complex models. The entire flow of our proposed method is shown in~\cref{algorithm2}.

    \renewcommand{\algorithmicrequire}{{\textbf{Input:}}}
    \renewcommand{\algorithmicensure}{{\textbf{Output:}}}
    \begin{algorithm}[ht]
    	\caption{Prototype Box Selection (PBR)}
	\begin{algorithmic}[1]
		\REQUIRE  The frozen trained model in $f_{\theta_t}(\cdot)$, the stream data $D^t$ at current task $t$, each image $I^t_n$ has $G_n^t$ groundtruth labels $\{y_g\}_{g=1}^{G_n^t}$, the box rehearsal memory $B^{t-1}$ after task $t-1$, the box rehearsal memory size $M$, the seen classes $\mathcal{C}^{1:t}$ until task $t$.
		\ENSURE The updated $B^{t}$ after task $t$. \\
		\STATE {\textbf{Initialize}}: $B^{t} = \{\}$, $m^t$ = ceil($M/|\mathcal{C}^{1:t}|$);
            \STATE $F^t_g$ = $f_{\theta_t}(I^t_n, y_g)$, $\forall{n} \in N^t$, $\forall{g} \in G^t_n$;
            \STATE $b_g=crop(I^t_n, y_g)$, 
            $\forall{n} \in N^t$, $\forall{g} \in G^t_n$;      
        \FOR {$c$ in $\mathcal{C}^{1:t}$}
        \IF{$c \in \mathcal{C}^{t}$}
            \STATE Compute $\hat{F}^{t}_{c}$ for each class $c$ based on~\cref{eq:profm};
            \STATE $D_c = \left\{(b_g, y_g)\mid c_g=c\right\}$;
            \STATE Sort ${D}_c$ following~\cref{eq:dis};
            \STATE $B^{t}+=D_c[0:m^t]$;
        \ELSE
            \FOR {$j=1, 2, ..., m^t$}
    		\STATE $i=j *\left|{B}_c^{t-1}\right| / ceil(M/|\mathcal{C}^{1:t-1}|)$;
    		\STATE ${B}^t+={B}_c^{t-1}[i]$;
    		\ENDFOR
       \ENDIF\ENDFOR
	\end{algorithmic}
	\label{alg:1}
    \end{algorithm}

    \renewcommand{\algorithmicrequire}{{\textbf{Input:}}}
    \renewcommand{\algorithmicensure}{{\textbf{Output:}}}
    \begin{algorithm}[ht]
    	\caption{Augmented Box Replay Method}
    	\begin{algorithmic}[1]
    		\REQUIRE  $f_{\theta_{t-1}}(\cdot)$, $D^t$=$\{I^t_n, G_n^t\}_{n=1}^{N_t}$, $B^{t-1}$ and Rat=1:1:2.
    		\ENSURE The updated $B^{t}$ and $f_{\theta_t}(\cdot)$ after task $t$. \\
    		\STATE {\textbf{Initialize}}: $\theta_{t} = \theta_{t-1}$;
            \FOR {$n$ in $N_t$}
                \STATE MIX,MOS,NEW=GenerateReplayType(Rat);
                \IF{MIX}
                    \STATE Compute $\hat{I}^t_n, \hat{G}^t_n$ by MixupBoxReply($I^t_n, G_n^t$);
                \ELSIF{MOS}
                    \STATE Compute $\hat{I}^t_n, \hat{G}^t_n$ by MosaicBoxReply($I^t_n, G_n^t$);
                \ELSIF{NEW}
                    \STATE $\{\hat{I}^t_n, \hat{G}^t_n\} = \{I^t_n, G_n^t$\};
                \ENDIF
                \STATE $\mathcal{L}_{Dis}=$ DistiallationLosses($f_{\theta_{t-1}}(\cdot), f_{\theta_{t}}(\cdot), \hat{I}^t_n$);
                \STATE $\mathcal{L}_{Det}=$ DetectionLosses($f_{\theta_{t}}(\cdot), \{\hat{I}^t_n, \hat{G}^t_n\}$);
                \STATE Update $\theta_{t}$ by $\mathcal{L}_{Dis}+\mathcal{L}_{Det}$;
           \ENDFOR
           \STATE Update $B_t$ by PBS($f_{\theta_{t}}(\cdot), D^t, B^{t-1}$);
    	\end{algorithmic}
    	\label{algorithm2}
    \end{algorithm}

\section{Additional Analysis}
\subsection{Analysis foreground shift problem}

In Table \textcolor{red}{1} and Table \textcolor{red}{2}, our algorithm demonstrates a remarkable improvement in mean Average Precision (mAP) ranging from 0.2$\sim$20\% across all categories. Additionally, it exhibits a substantial mAP boost of 4.5\% to 25.2\% in new categories (foreground categories), indicating the enhanced stability and plasticity achieved by our method.

\begin{figure}[t]
  \centering
    \includegraphics[width=\linewidth,height=0.7\linewidth]{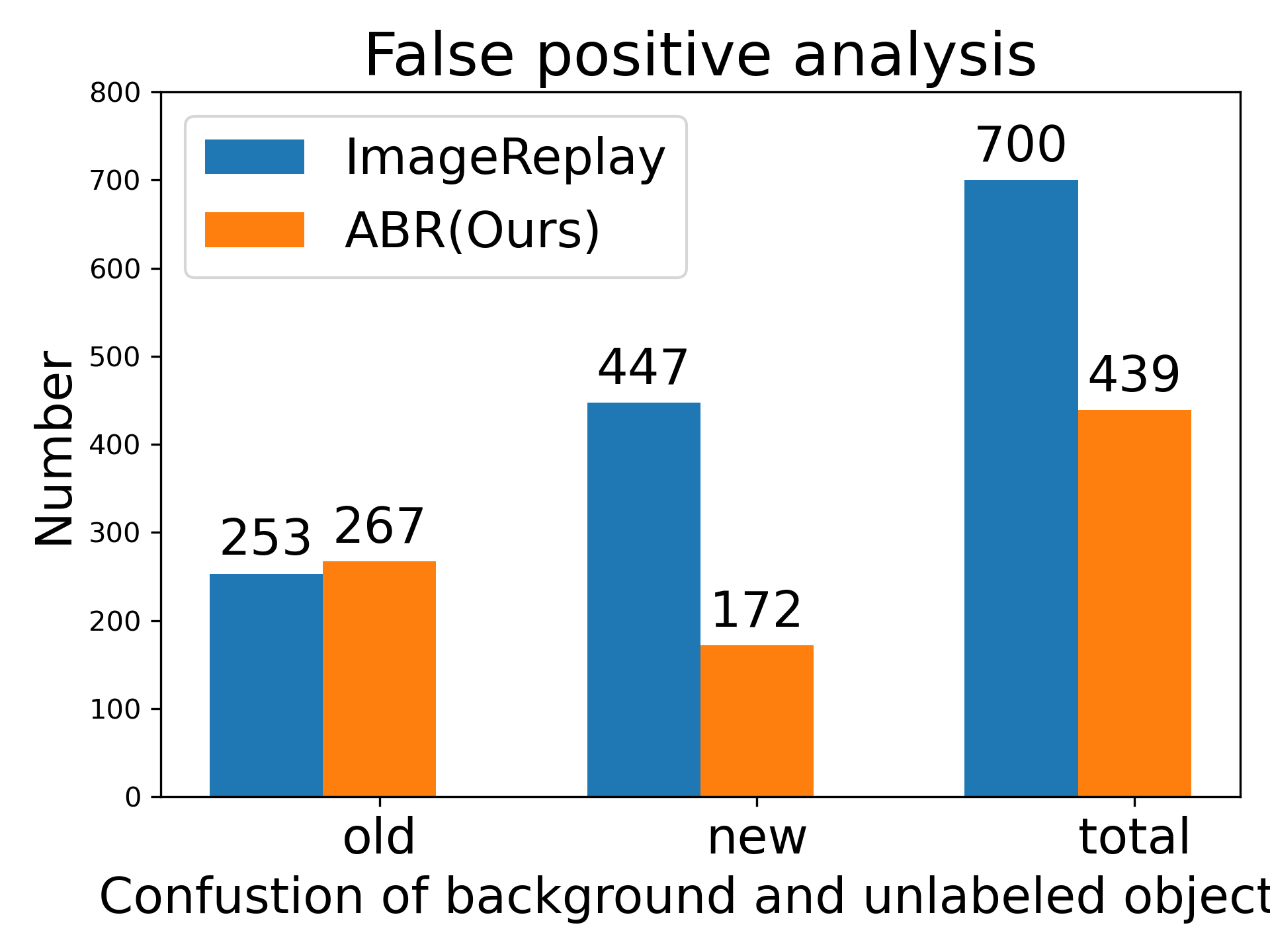}
    \caption{False-Positive Analysis}
    \label{fig:sm-fp_bg}
    \end{figure}

Moreover, we conducted a comprehensive analysis of False Positives (FP)\cite{hoiem2012diagnosing} under the VOC 10-10 setting. \cref{fig:sm-fp_bg} visually represents the number of background errors, specifically detections confused with the background or unlabeled objects. Notably, our approach (ABR) demonstrates a clear advantage, exhibiting a substantial reduction of 275 errors in new (foreground) classes compared to the ImageReplay method. This compelling result strongly suggests the successful mitigation of the foreground shift problem by our proposed approach.

\subsection{Analysis Attentive RoI Distillation (ARD)}

While existing methods have utilized attention distillation primarily on feature maps, we advance this approach by integrating location information of Region of Interest (RoI) proposals. By doing so, our model gains the capability to distill both feature and localization information from the replayed and new objects, leading to an overall performance enhancement.

\begin{figure}[t]
  \centering  
  \begin{subfigure}[t]{0.322\linewidth}
    \includegraphics[width=\linewidth,height=0.7\linewidth]{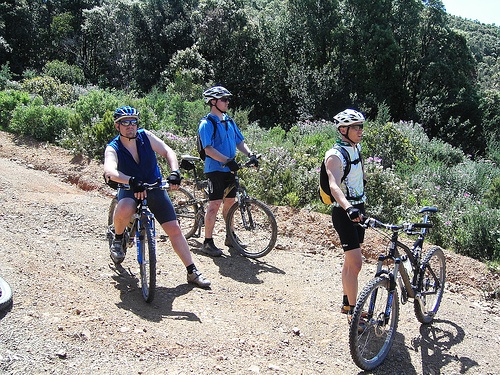}\vspace{-6pt}
    \caption{Image}
  \end{subfigure}
  \begin{subfigure}[t]{0.322\linewidth}
    \includegraphics[width=\linewidth,height=0.7\linewidth]{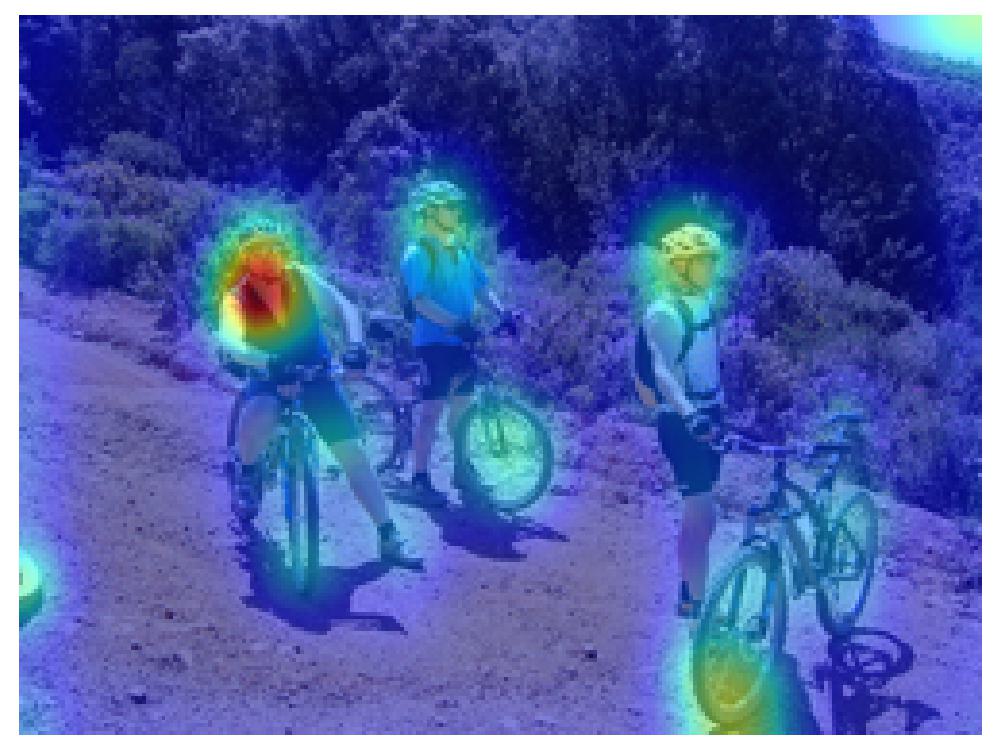}\vspace{-6pt}
    \caption{w/o ARD}
  \end{subfigure}
  \begin{subfigure}[t]{0.322\linewidth}
    \includegraphics[width=\linewidth,height=0.7\linewidth]{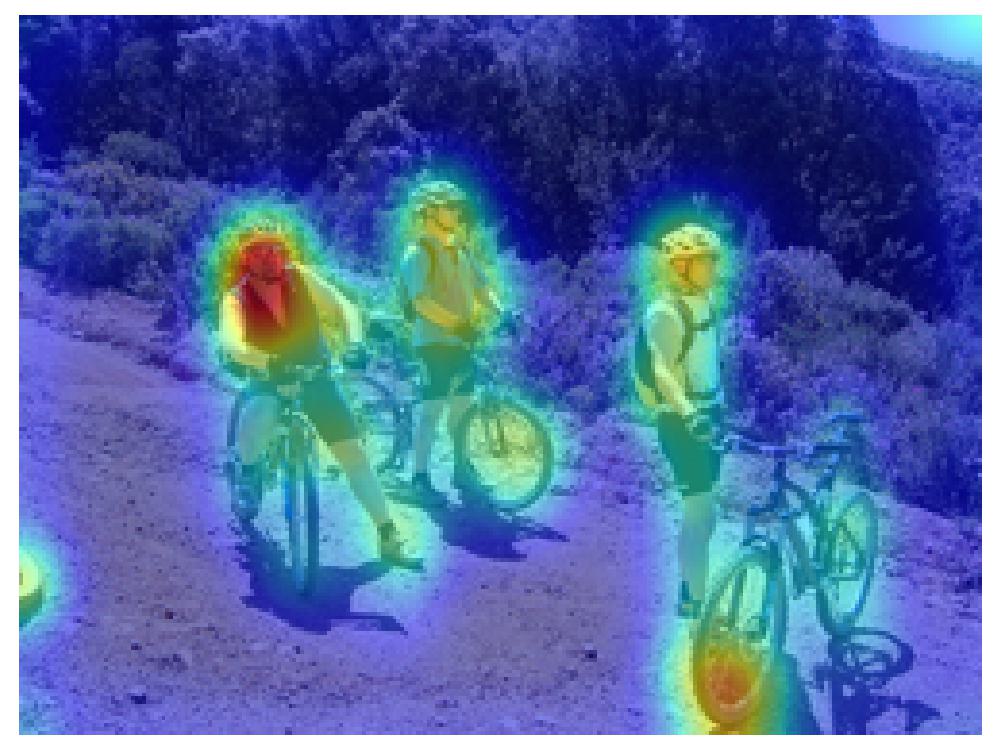}\vspace{-6pt}
    \caption{w ARD}
  \end{subfigure}
  \caption{Attention maps during training (person and bicycle are new and old classes respectively).}\label{fig:sm-attmap}
  \end{figure}

\cref{fig:sm-attmap} showcases some additional attention maps, highlighting how our Attention-based RoI Distillation (ARD) loss effectively retains attention on the old class (e.g., bicycle). This observation confirms ARD's competence in alleviating catastrophic forgetting, a phenomenon that impacts model performance when learning new tasks.

Through the inclusion of location-awareness in attention distillation, our proposed ARD method exemplifies its potential to mitigate catastrophic forgetting and reinforce the preservation of crucial knowledge from previous tasks, resulting in improved overall model performance.

\subsection{Effect of Hyperparameters}
\begin{figure}[t]
  \centering
    \includegraphics[width=0.49\linewidth,height=0.35\linewidth]{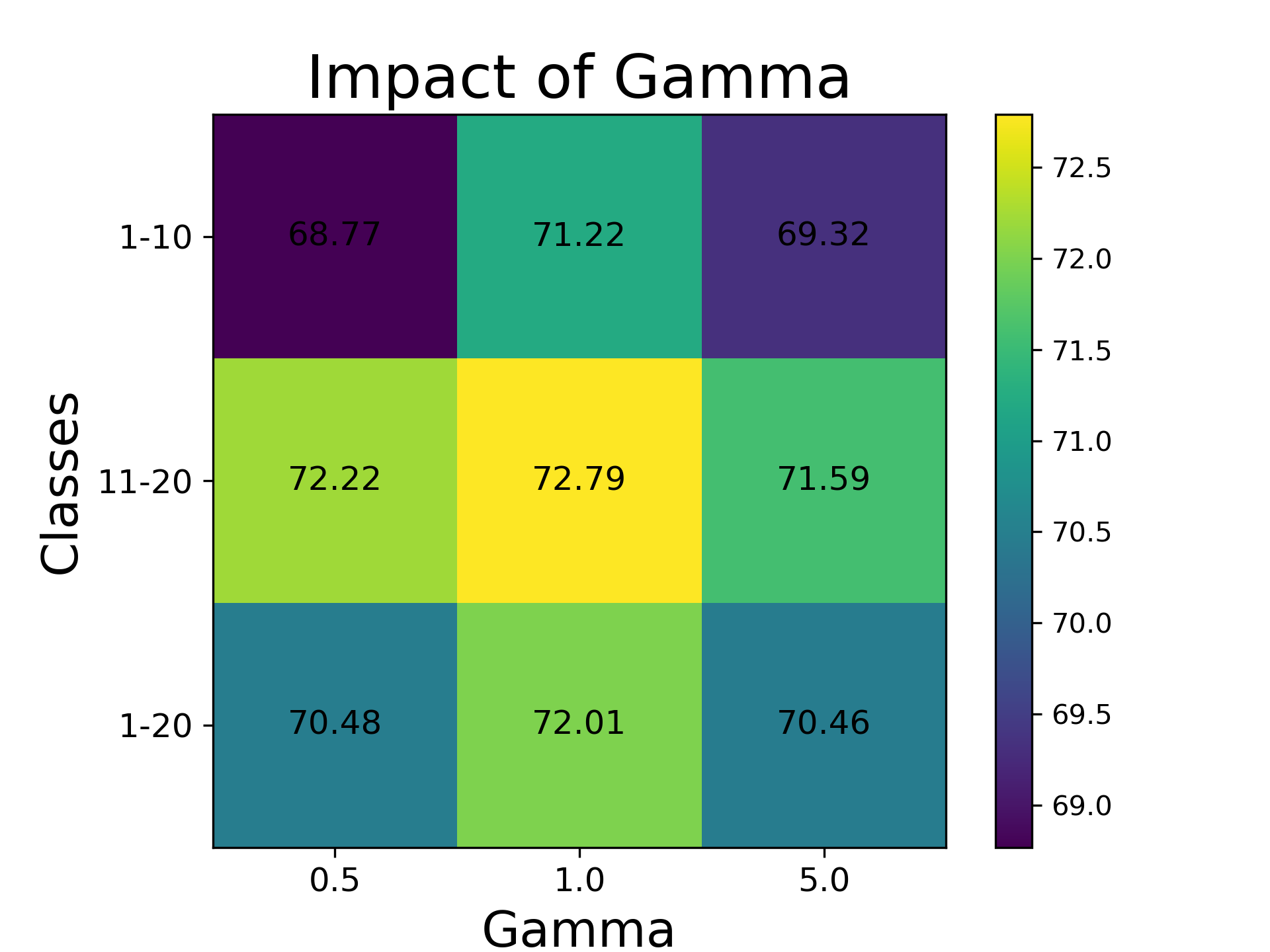}
    \includegraphics[width=0.49\linewidth,height=0.35\linewidth]{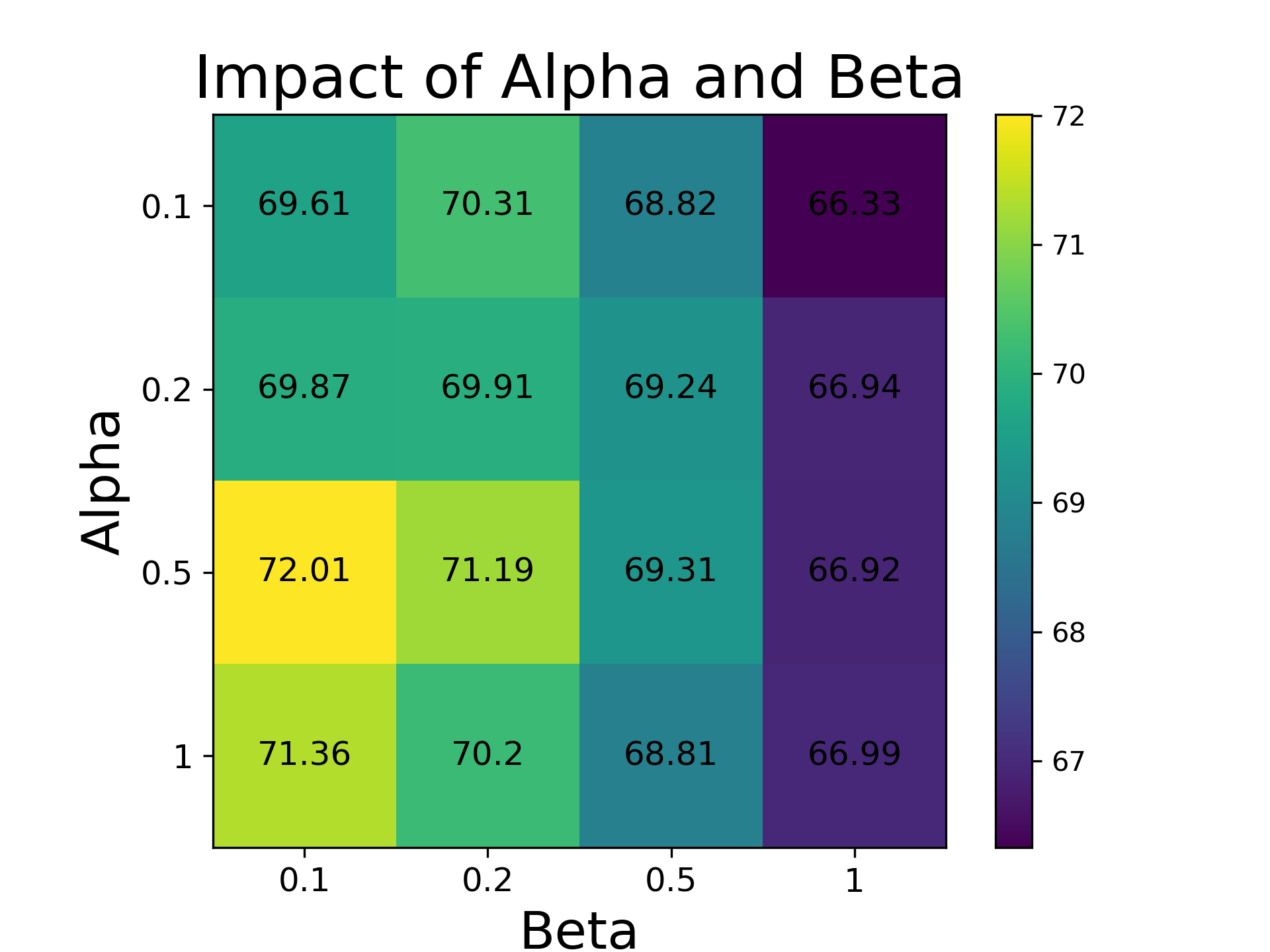}
  \caption{Impact of the hyperparameters $\gamma$, $\alpha$ and $\beta$.}
    \label{fig:sm-hyp}
 \end{figure}
We conducted additional experiments under the VOC 10-10 setting to analyze the impact of all hyperparameters in our study, as depicted in~\cref{fig:sm-hyp}. 
For $\gamma$ in Eq. \textcolor{red}{5} of the overall ARD loss function, we vary it in range [0.5, 1.0, 5.0]. From the results shown in the first figure of~\cref{fig:sm-hyp}, we find that the default $\gamma=1$ provides good results. 

In consequence, we optimize the total objective function to realize incremental object detecion learning:
\begin{equation}
    \mathcal{L}_{total} = \mathcal{L}_{faster\_rcnn} + \alpha\mathcal{L}_{ID}+\beta\mathcal{L}_{ARD}
\end{equation}
where $\alpha$ and $\beta$ weight for the Inclusive Distillation Loss and Attentive RoI Distillation, respectively. We vary it in range [0.1, 0.2, 0.5, 1].
The performance varies as a function of $\alpha, \beta$ outperforming the state-of-the-art (66.8) for most combinations.

\section{Additional Results}\label{sec:ar}
\subsection{Detailed Results for the Long Sequences}
\begin{table*}[ht]
\centering
\caption{Per-Class AP@50 and Overall mAP@50 values in different task on PASCAL-VOC 2007 5-5 setting. }
\label{tab:longseq}
\renewcommand\arraystretch{1.2}
\scriptsize

\setlength{\tabcolsep}{0.5pt}
\begin{tabular}{l|l|cccccccccc:c:ccccc:c:ccccc:c|c}
\multicolumn{1}{l|}{\textbf{Class Split}} & \textbf{Method} & \textbf{aero} & \textbf{cycle} & \textbf{bird} & \textbf{boat} & \textbf{bottle} & \textbf{bus} & \textbf{car} & \textbf{cat} & \textbf{chair} & \textbf{cow} & \multicolumn{1}{c:}{\underline{\textbf{mAP-task1}}} & \textbf{table} & \textbf{dog} & \textbf{horse} & \textbf{bike} & \textbf{person} & \multicolumn{1}{c:}{\underline{\textbf{mAP-task2}}} & \multicolumn{1}{c}{\textbf{plant}} & \multicolumn{1}{c}{\textbf{sheep}} & \multicolumn{1}{c}{\textbf{sofa}} & \multicolumn{1}{c}{\textbf{train}} & \multicolumn{1}{c:}{\textbf{tv}} & \multicolumn{1}{c|}{\underline{\textbf{mAP-task3}}} & \multicolumn{1}{c}{\underline{\textbf{mAP-total}}} \\ \hline\hline
\textbf{1-20} & \textbf{JT} & 72.7  & 81.0  & 76.0  & 58.9  & 62.0  & 76.4  & 87.4  & 85.7  & 72.6  & 82.4  & 75.5  & 57.7  & 83.2  & 85.7  & 80.5  & 84.2  & 78.3  & \multicolumn{1}{c}{45.8 } & \multicolumn{1}{c}{77.1 } & \multicolumn{1}{c}{65.9 } & \multicolumn{1}{c}{75.7 } & \multicolumn{1}{c:}{74.5 } & 67.8  & 74.3  \\\hline
\multirow{2}{*}{\textbf{(1-5)+6-10}} & \textbf{MMA} & 73.8  & 80.8  & 71.2  & 52.5  & 63.3  & 55.2  & 74.9  & 65.2  & 39.1  & 73.3  & 64.9  & \multicolumn{1}{c}{} & \multicolumn{1}{c}{} & \multicolumn{1}{c}{} & \multicolumn{1}{c}{} & \multicolumn{1}{c:}{} &  & \multicolumn{1}{c}{} & \multicolumn{1}{c}{} & \multicolumn{1}{c}{} & \multicolumn{1}{c}{} & \multicolumn{1}{c:}{}  &  & 64.9  \\
 & \textbf{ABR} & 71.7  & 82.6  & 69.5  & 53.6  & 63.8  & 63.0  & 79.0  & 68.5  & 47.0  & 78.4  & \textbf{67.7}  &  &  &  &  &  &  &  &  &  &  &  &  & \textbf{67.7}  \\\hline
\multirow{2}{*}{\textbf{(1-10)+11-15}} & \textbf{MMA} & 67.4  & 78.1  & 64.5  & 49.7  & 63.5  & 23.1  & 34.5  & 26.3  & 8.7  & 35.0  & 45.1  & 47.5  & 52.8  & 67.5  & 65.9  & 76.0  & 61.9  &  &  &  &  &  &  & 50.7  \\
 & \textbf{ABR} & 68.5  & 79.6  & 67.3  & 51.9  & 56.7  & 60.2  & 75.2  & 62.8  & 38.6  & 62.0  & \textbf{62.3}  & 54.0  & 66.3  & 76.9  & 74.5  & 77.3  & \textbf{69.8}  & \multicolumn{1}{c}{} & \multicolumn{1}{c}{} & \multicolumn{1}{c}{} & \multicolumn{1}{c}{} & \multicolumn{1}{c:}{} &  & \textbf{64.8}  \\\hline
\multirow{2}{*}{\textbf{(1-15)+16-20}} & \textbf{MMA} & 72.3  & 75.5  & 57.0  & 46.9  & 59.9  & 4.8  & 32.4  & 38.5  & 3.3  & 1.4  & 39.2  & 0.7  & 28.8  & 42.2  & 44.1  & 18.2  & 26.8  & \multicolumn{1}{c}{36.0 } & \multicolumn{1}{c}{46.5 } & \multicolumn{1}{c}{52.0 } & \multicolumn{1}{c}{52.0 } & \multicolumn{1}{c:}{66.6 } & 50.6  & 38.9  \\
 & \textbf{ABR} & 69.3  & 80.0  & 65.6  & 53.9  & 54.6  & 52.2  & 75.5  & 69.4  & 34.3  & 69.6  & \textbf{62.4}  & 22.9  & 41.8  & 48.7  & 53.7  & 60.8  & \textbf{45.6}  & \multicolumn{1}{c}{39.6 } & \multicolumn{1}{c}{71.3 } & \multicolumn{1}{c}{59.2 } & \multicolumn{1}{c}{76.1 } & \multicolumn{1}{c:}{70.4 } & \textbf{63.3}  & \textbf{58.4}
\end{tabular}
\end{table*}
In \cref{tab:longseq}, we present the results of our experiments with long sequences on the PASCAL-VOC 2007 dataset. To simulate this scenario, we trained our detector on images from the first 5 classes and gradually added classes 6 to 20 in groups of five.

The table shows the class-wise average precision (AP)@0.5 and the corresponding mean average precision (mAP). The first row (JT) represents the upper-bound where the detector is trained on data from all 20 classes. The subsequent three pairs of rows demonstrate the results obtained when adding five new classes at a time. The notation (1-5)+6..10 is used to represent this setting. Our proposed ABR method outperforms the previous state-of-the-art method MMA~\cite{cermelli2022modeling} on all sequential tasks, as can be seen from the results in \cref{tab:longseq}. Therefore, the ABR method can be more useful in real-world scenarios where new object classes are frequently introduced. Additionally, the ABR method is a novel approach that may have implications for future research in object detection.

\subsection{Visualization}
\begin{figure*}[htbp]
  \begin{subfigure}[b]{0.24\linewidth}
    \includegraphics[width=\linewidth,height=0.7\linewidth]{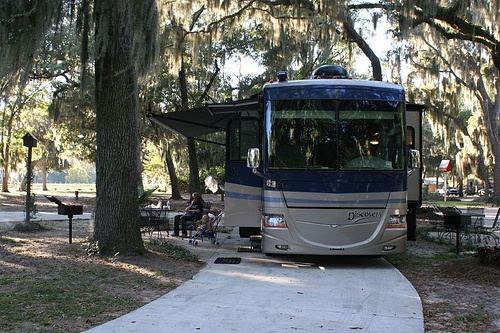}
    \includegraphics[width=\linewidth,height=0.7\linewidth]{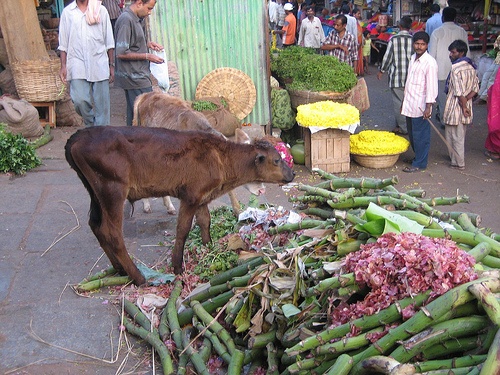}
    \includegraphics[width=\linewidth,height=0.7\linewidth]{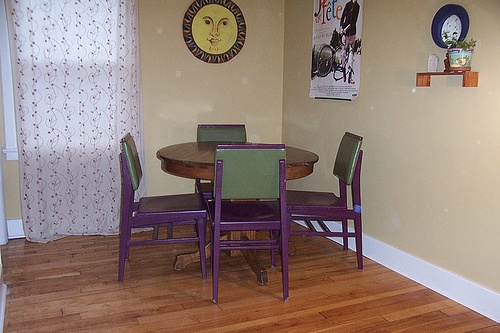}
    \includegraphics[width=\linewidth,height=0.7\linewidth]{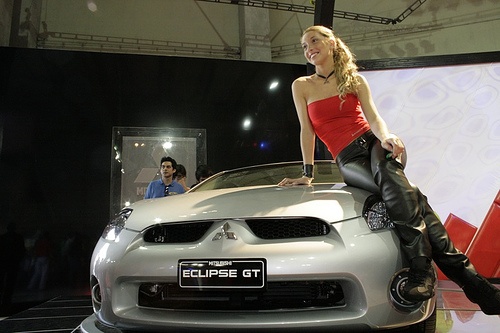}
    \includegraphics[width=\linewidth,height=0.7\linewidth]{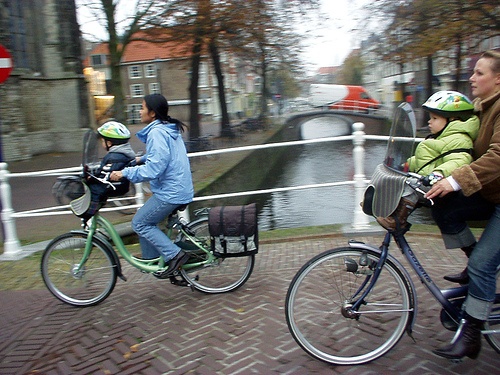}
    \includegraphics[width=\linewidth,height=0.7\linewidth]{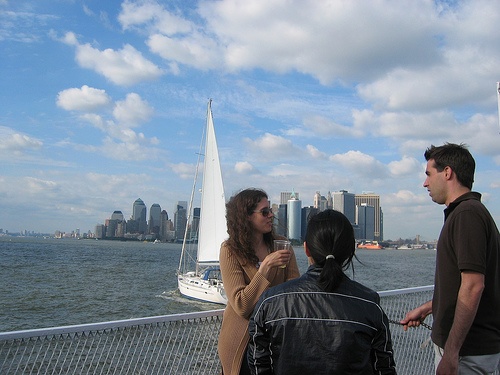}
    \includegraphics[width=\linewidth,height=0.7\linewidth]{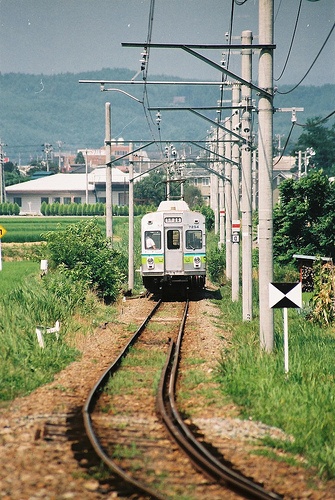}
    \caption{Image}
  \end{subfigure}
  \begin{subfigure}[b]{0.24\linewidth}
    
    \includegraphics[width=\linewidth,height=0.7\linewidth]{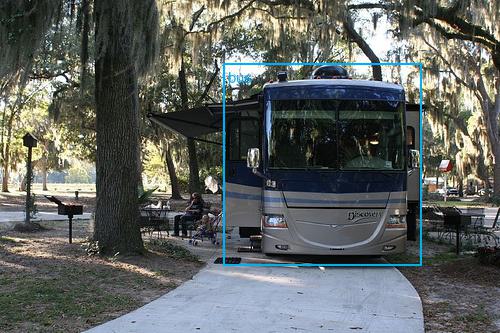}
    \includegraphics[width=\linewidth,height=0.7\linewidth]{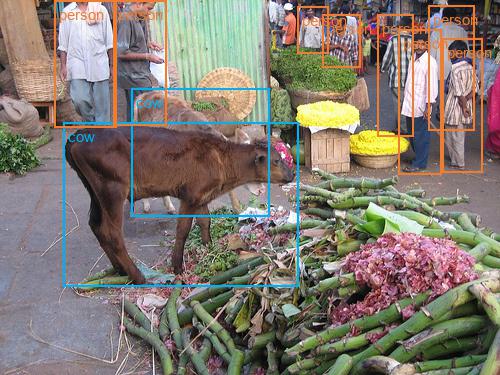}
    \includegraphics[width=\linewidth,height=0.7\linewidth]{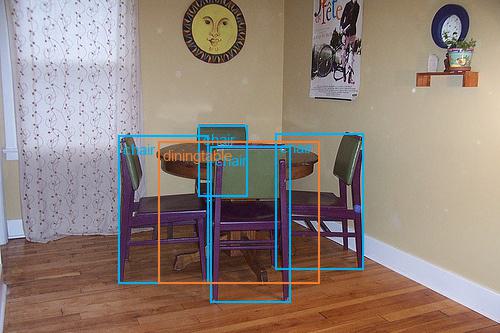}
    \includegraphics[width=\linewidth,height=0.7\linewidth]{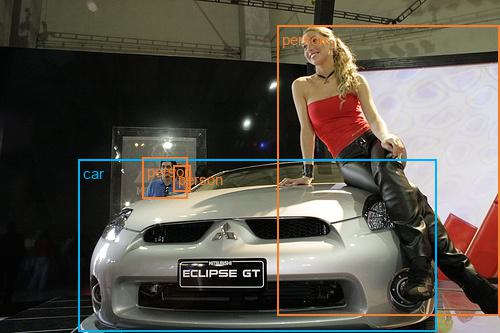}
    \includegraphics[width=\linewidth,height=0.7\linewidth]{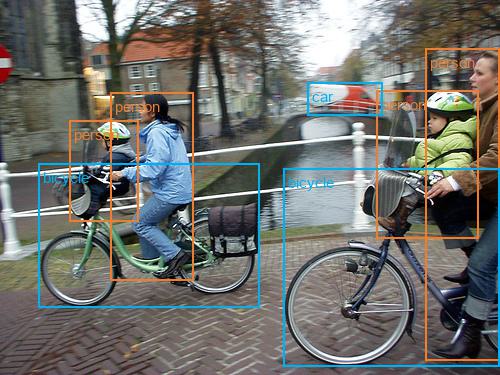}
    \includegraphics[width=\linewidth,height=0.7\linewidth]{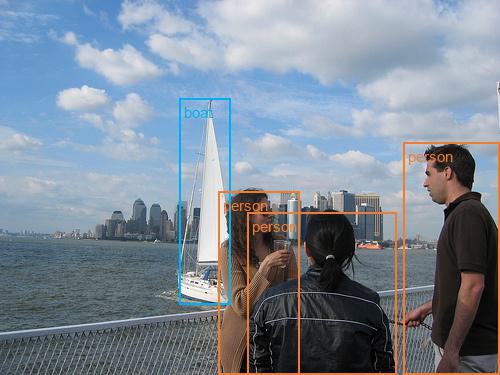}
    \includegraphics[width=\linewidth,height=0.7\linewidth]{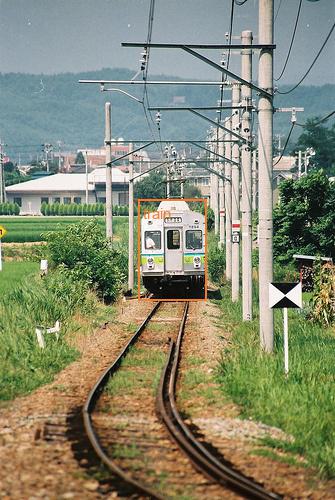}
    \caption{GT}
  \end{subfigure}
  \begin{subfigure}[b]{0.24\linewidth}
    
    \includegraphics[width=\linewidth,height=0.7\linewidth]{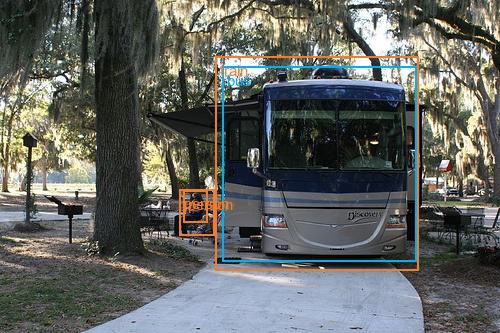}
    \includegraphics[width=\linewidth,height=0.7\linewidth]{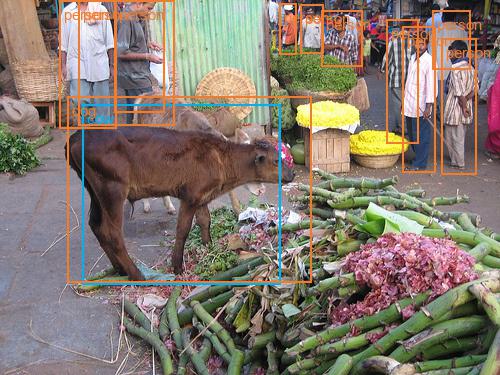}
    \includegraphics[width=\linewidth,height=0.7\linewidth]{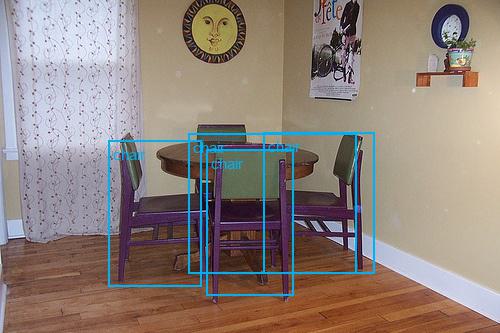}
    \includegraphics[width=\linewidth,height=0.7\linewidth]{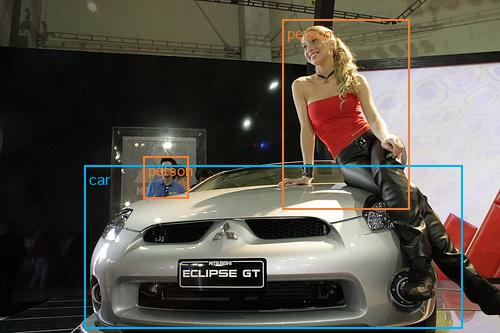}
    \includegraphics[width=\linewidth,height=0.7\linewidth]{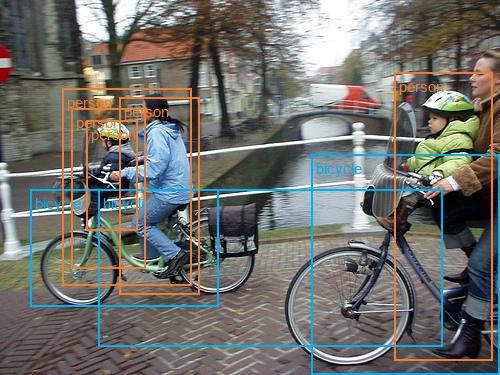}
    \includegraphics[width=\linewidth,height=0.7\linewidth]{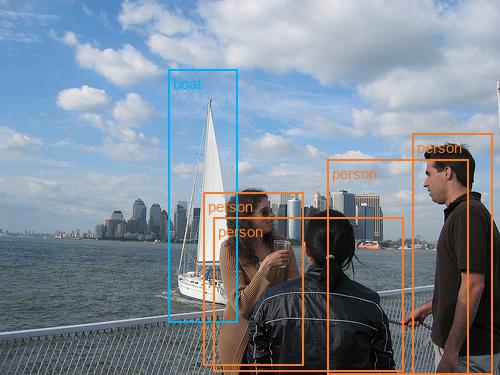}
    \includegraphics[width=\linewidth,height=0.7\linewidth]{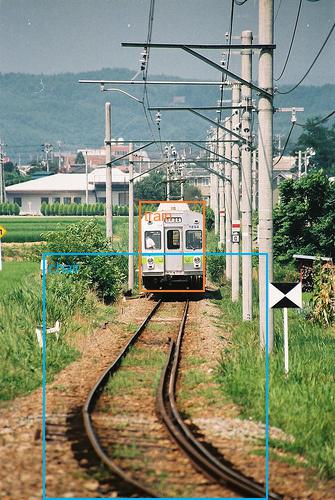}
    \caption{MMA}
  \end{subfigure}
  \begin{subfigure}[b]{0.24\linewidth}
    
    \includegraphics[width=\linewidth,height=0.7\linewidth]{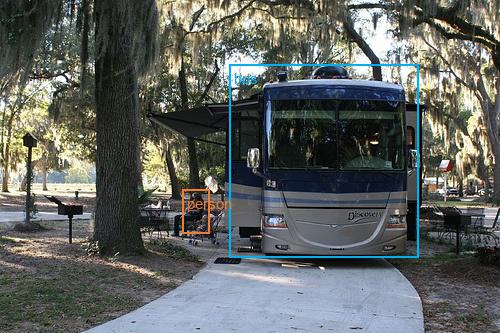}
    \includegraphics[width=\linewidth,height=0.7\linewidth]{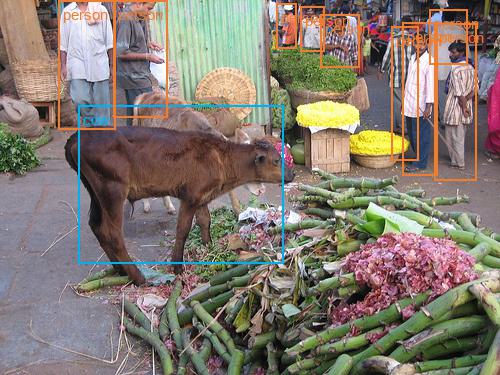}
    \includegraphics[width=\linewidth,height=0.7\linewidth]{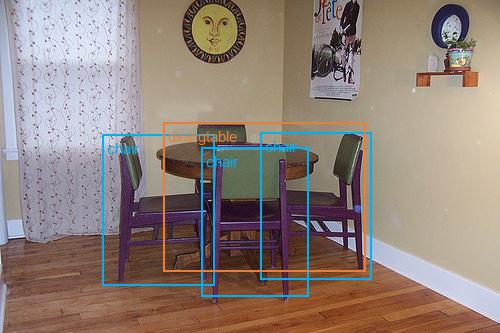}
    \includegraphics[width=\linewidth,height=0.7\linewidth]{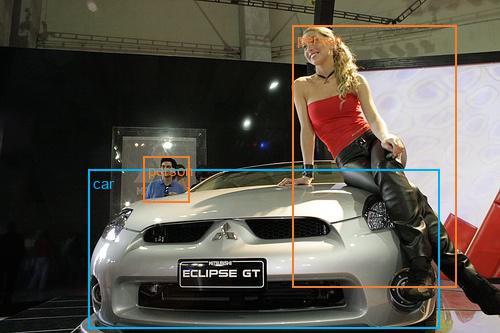}
    \includegraphics[width=\linewidth,height=0.7\linewidth]{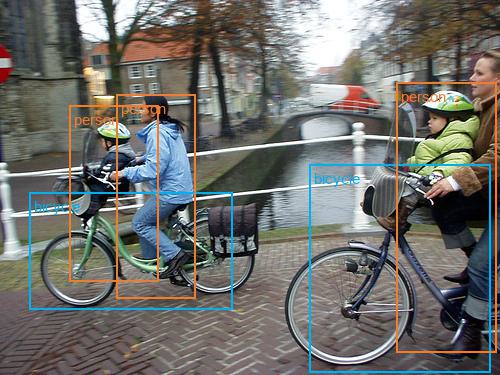}
    \includegraphics[width=\linewidth,height=0.7\linewidth]{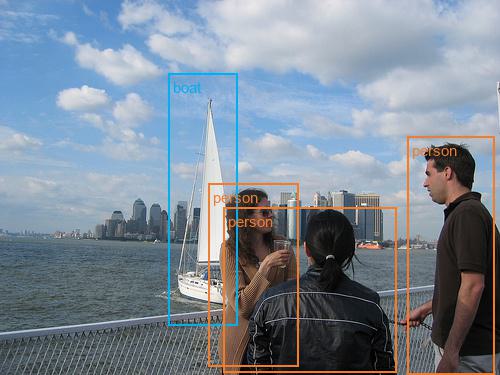}
    \includegraphics[width=\linewidth,height=0.7\linewidth]{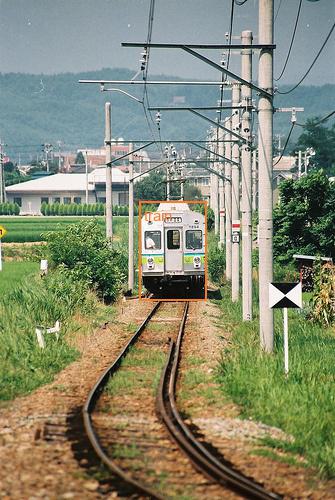}
    \caption{Ours}
  \end{subfigure}
  \caption{Visualization of the inference results in MMA and Ours for 8 test images on PASCAL-VOC 2007 10-10 scenario.}
    \label{fig:addpred}
\end{figure*}

The inference results are presented in \cref{fig:addpred}, which demonstrate the effectiveness of our proposed ABR method in avoiding the forgetting of previous classes and improving adaptation to new classes. 
In the first two rows, our method is capable of accurately distinguishing new classes from similar classes in the previous classes, as seen in the detection of a \textit{bus} in the first row of images and a \textit{cow} in the second row of images. However, the popular MMA method misclassifies the \textit{bus} as a \textit{train} or \textit{bus} and the \textit{cow} as a \textit{dog} or \textit{cow}. In the third row, our algorithm successfully detects the new class, a \textit{dining table}, while also accurately locating a previous class, a \textit{chair}. In comparison to the MMA method, our method achieves more precise position detection, as demonstrated in the last two rows where \textit{person} and \textit{boat} are detected.

Overall, these results suggest that the proposed ABR method can more effectively handle the problem of incremental learning in object detection tasks, particularly in scenarios where new classes are similar to previous ones. The ability to avoid forgetting and adapt to new classes is crucial for practical applications, and the improved performance of our method is promising for future research in this area.

\nocite{hoiem2012diagnosing，cermelli2022modeling}
\end{document}